\documentclass[10pt,twocolumn,letterpaper]{article}

\usepackage[pagenumbers]{cvpr} 

\definecolor{cvprblue}{rgb}{0.21,0.49,0.74}
\usepackage[pagebackref,breaklinks,colorlinks,allcolors=cvprblue]{hyperref}
\usepackage{fontawesome5}

\usepackage{colortbl} 
\usepackage{rotating} 
\usepackage{multirow} 
\usepackage{adjustbox} 
\usepackage{booktabs}
\usepackage[table]{xcolor}
\usepackage{color}
\usepackage{caption}
\usepackage{subcaption}
\usepackage{makecell} 
\usepackage{tikz} 
\usepackage{array} 
\usepackage{float} 
\usepackage{afterpage}
\usepackage{placeins}
\usepackage{amssymb}
\usepackage{pifont}
\usepackage{graphicx}     

\definecolor{color1}{RGB}{176, 36, 24}
\definecolor{color2}{RGB}{119,185,0}
\definecolor{color3}{RGB}{0, 0, 200}

\def\paperID{*} 
\def\confName{CVPR}
\def\confYear{2026}

\title{ShelfGaussian: Shelf-Supervised Open-Vocabulary Gaussian-based \\ 3D Scene Understanding}

\author{
Lingjun Zhao\textsuperscript{*}\textsuperscript{\dag} \quad Yandong Luo\textsuperscript{*} \quad James Hays \quad Lu Gan\\
Georgia Institute of Technology, Atlanta, GA USA\\
\tt\small \{lzhao360,yluo471,hays,lgan\}@gatech.edu \\
}

\begin{document}
\maketitle

\let\thefootnote\relax\footnotetext{\textsuperscript{*}Equal contribution.}
\let\thefootnote\relax\footnotetext{\textsuperscript{\dag}Corresponding author.}

\begin{abstract}
We introduce \textbf{ShelfGaussian}, an open-vocabulary multi-modal Gaussian-based 3D scene understanding framework supervised by off-the-shelf vision foundation models (VFMs). Gaussian-based methods have demonstrated superior performance and computational efficiency across a wide range of scene understanding tasks. However, existing methods either model objects as closed-set semantic Gaussians supervised by annotated 3D labels, neglecting their rendering ability, or learn open-set Gaussian representations via purely 2D self-supervision, leading to degraded geometry and limited to camera-only settings. To fully exploit the potential of Gaussians, we propose a Multi-Modal Gaussian Transformer that enables Gaussians to query features from diverse sensor modalities, and a Shelf-Supervised Learning Paradigm that efficiently optimizes Gaussians with VFM features jointly at 2D image and 3D scene levels. We evaluate ShelfGaussian on various perception and planning tasks. Experiments on Occ3D-nuScenes demonstrate its state-of-the-art zero-shot semantic occupancy prediction performance. ShelfGaussian is further evaluated on an unmanned ground vehicle (UGV) to assess its in-the-wild performance across diverse urban scenarios. Project website: \href{https://lunarlab-gatech.github.io/ShelfGaussian/}{https://lunarlab-gatech.github.io/ShelfGaussian/}.
\end{abstract}
\vspace{-15pt}

\section{Introduction}
\label{sec:intro}
3D scene understanding plays a crucial role in autonomous driving and robotic navigation. Among various tasks, 3D semantic occupancy prediction is particularly challenging, as it requires a fine-grained, joint semantic and geometric understanding of the surroundings~\cite{openoccupancy, occ3d, occnet}. Traditional semantic occupancy frameworks~\cite{tpvformer, occformer, voxformer, surroundocc} employ a dense voxel-based scene representation that is computationally and memory-intensive and unsuitable for on-board real-time deployment. Recently, 3D Gaussian Splatting (3DGS)~\cite{3dgs} is naturally extended into an efficient and expressive scene representation for occupancy prediction, leveraging its object-centric modeling and high-fidelity rendering capabilities. Some work \cite{gaussianocc, gaussrender, gaussianpretrain} retains voxels as the scene representation and use Gaussian rendering to improve feature learning. However, they inherit the fundamental limitations of voxel-based representations. In contrast, GaussianFormer family~\cite{gaussianformer, gaussianformer2, gaussianworld, gaussianformer3d, graphgsocc} directly adopts Gaussians as an alternative to voxels but still trains Gaussians to predict predefined semantics in a fully-supervised manner, which is limited to closed-set scenarios.

\begin{figure}[t] 
    \centering
    \includegraphics[width=\columnwidth]{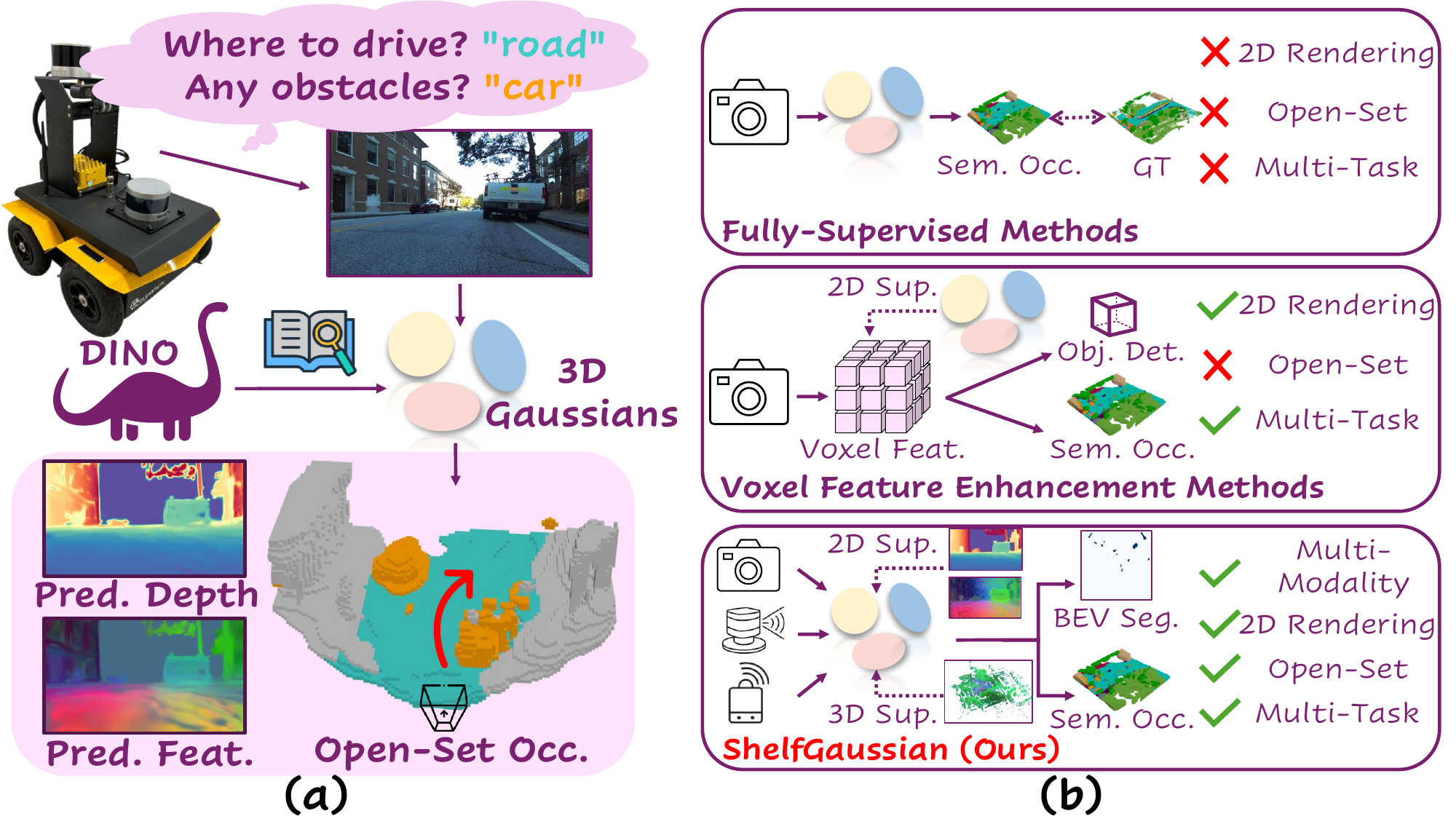}
    \vspace{-18pt}
    \caption{We propose \textbf{ShelfGaussian} for Gaussian-based 3D scene understanding under \textit{open-vocabulary}, \textit{multi-modal} and \textit{multi-task} scenario. \textbf{(a)} Our model is able to assist a robot in predicting open-set occupancy from any sensor modalities with the help of VFMs. \textbf{(b)} Compared to existing Gaussian-based methods, ours provides a generalizable solution for 3D scene understanding.}
    \vspace{-10pt}
    \label{fig:teaser}
\end{figure}

This closed-vocabulary constraint is a bottleneck for real-world deployment. Autonomous vehicles inevitably encounter semantics beyond a predefined set and lead to unpredictable behaviors. The reliance on fully-supervised training also impedes model's ability to scale with data. These challenges have spurred a significant push towards self-supervised, open-vocabulary 3D scene understanding. VFMs such as CLIP~\cite{clip} and DINO~\cite{dino, dinov2, dinov3}, demonstrate superior performance in open-set 2D visual perception, making them key enablers for extending this success to 3D. 
However, effectively leveraging 2D VFMs for 3D scene understanding remains challenging due to fundamental data and representation discrepancies between domains.

Recent attempts, such as GaussTR~\cite{gausstr} and GaussianFlowOcc~\cite{gaussianflowocc}, learn feed-forward open-vocabulary Gaussians from camera-only information by using outputs from VFMs for 2D self-supervision. While these methods established a VFM-enabled self-supervised pipeline for 3D occupancy prediction, a significant performance gap remains compared to fully-supervised ones. These observations motivate us to explore stronger supervision signals and complementary input modalities beyond 2D images to fully unlock the expressiveness of Gaussian-based scene representations for open-vocabulary 3D scene understanding.

However, effectively learning open-vocabulary Gaussian features directly from unannotated 3D data is non-trivial. One bottleneck is the lack of an efficient,  general Gaussian-to-voxel splatting module to support high-dimensional feature gradient backpropagation from voxel-based supervision. Another limitation is the absence of a systematic mechanism to bridge Gaussians with data from different 3D input modalities.
Driven by these opportunities and challenges, we dig deeper into the application of feed-forward Gaussians for 3D scene understanding under \textit{multi-modal}, \textit{open-vocabulary}, and \textit{multi-task} settings, and propose a \textbf{ShelfGaussian} framework. Our contributions are:
\begin{itemize}
    \item We propose a \textbf{multi-modal Gaussian transformer} architecture that queries features from different sensor modalities to predict feed-forward 3D Gaussians, supporting camera-only, LiDAR-only, LiDAR-camera, camera-radar and LiDAR-camera-radar sensor input.
    \item We introduce a novel \textbf{shelf-supervised learning paradigm} to optimize Gaussians using off-the-shelf VFMs at both 2D image and 3D scene levels. A highly efficient, CUDA-accelerated Gaussian-to-Voxel (G2V) splatting module is designed to enable high-dimensional VFM feature distillation and speed up training.
    \item ShelfGaussian achieves \textbf{state-of-the-art zero-shot performance} on semantic occupancy prediction on nuScenes dataset, superior performance on Bird's-Eye-View (BEV) segmentation, and reduced collision rate in trajectory planning when integrated into our Gaussian-Planner.
    \item We further test ShelfGaussian on an unmanned ground vehicle (UGV) across diverse urban scenes, demonstrating its \textbf{superior in-the-wild performance}. All code and data will be open-sourced to benefit the community.
\end{itemize}

\section{Related Work}
\label{sec:related_work}

\subsection{Open-Vocabulary 3D Scene Understanding}
Open-vocabulary 3D scene understanding typically includes 3D semantic segmentation~\cite{openscene, dma, clip2scene, pla, regionplc, lowis3d, sal}, 3D instance segmentation~\cite{unim-ov3d, openins3d, openmask3d, sal, open3dsg, search3d, open3dis, sam3d, sampro3d, sai3d, ovir-3d, maskclustering, opensu3d, octree-graph, sam4d}, 3D object classification~\cite{pointclip, pointclip-v2, clip2point, openins3d}, and 3D object detection~\cite{pointclip-v2, openins3d, ov-3det, shelf}.
For outdoor driving scenarios, open-vocabulary occupancy prediction is particularly prominent due to its fine-grained voxel-level output. OVO~\cite{ovo}, Veon~\cite{veon} and POP-3D~\cite{pop-3d} are pioneers in distilling CLIP~\cite{clip} features into 3D voxels to enable open-set querying abilities. 
Building on the success of 2D self-supervision in semantic occupancy prediction~\cite{selfocc, renderocc, simpleocc}, recent works such as OccNeRF~\cite{occnerf}, DistillNeRF~\cite{distillnerf}, and LangOcc~\cite{langocc} leverage volume rendering to align voxel features with foundation models, thereby enabling open-set occupancy prediction.
CAL~\cite{cal} and LOcc~\cite{locc} propose a CLIP-based 3D pseudo-labeling engine and leverage it to supervise occupancy networks.

\subsection{3D Gaussians for Occupancy Prediction}
3D Gaussian Splatting~\cite{3dgs} demonstrates superior reconstruction and rendering quality, and has been extended to semantic occupancy prediction. GaussianFormer~\cite{gaussianformer, gaussianformer2} models scenes as dense semantic Gaussians supervised by closed-set voxel-wise labels. This representation is further extended by GaussianWorld~\cite{gaussianworld} to temporal fusion and by GaussianFormer3D~\cite{gaussianformer3d} to multi-modality scenarios, respectively. GaussianOcc~\cite{gaussianocc}, GaussRender~\cite{gaussrender}, and GaussianPretrain~\cite{gaussianpretrain} all employ a voxel-based representation and enhance voxel features through Gaussian rendering. GaussTR~\cite{gausstr} learns sparse 3D Gaussians as the scene representation by aligning Gaussian renderings with 2D vision foundation features. GaussianFlowOcc~\cite{gaussianflowocc} further incorporates a temporal module for Gaussians to model scene dynamics, achieving improved performance.
To the best of our knowledge, our proposed ShelfGaussian is the first framework that unifies \textit{open-vocabulary}, \textit{multi-modality} and \textit{Gaussian-based} scene completion.


\section{Method}
\label{sec:method}

\begin{figure*}[htp] 
    \centering
    \includegraphics[height=7cm]{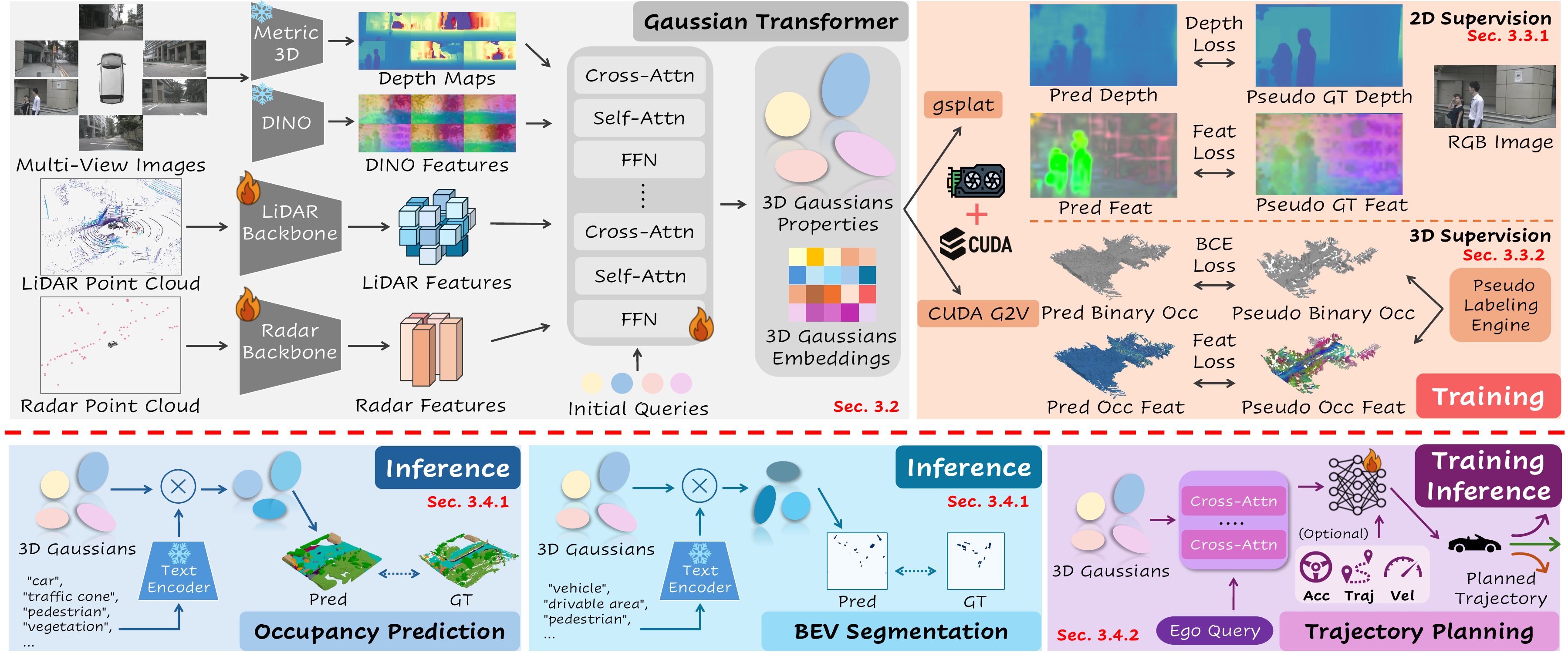}
    \vspace{-5pt}
    \caption{\textbf{Overview of ShelfGaussian.} 
   ShelfGaussian employs off-the-shelf VFMs to extract depth and DINO feature maps from multi-view images, and trains LiDAR and radar backbones to extract related features. These are then fed into our multi-modal Gaussian transformer to predict sparse sets of 3D Gaussians to represent the scene. During training, Gaussians are rendered into camera views for VFM-based 2D supervision, while being converted into voxels via our CUDA-accelerated G2V module for 3D supervision. The shelf-supervised Gaussians support zero-shot semantic occupancy prediction, BEV segmentation, and are further evaluated for trajectory planning. }
    \vspace{-8pt}
    \label{fig:main_method}
\end{figure*}

This section detailed the proposed ShelfGaussian framework, with an overview illustrated in~\cref{fig:main_method}.

\subsection{Preliminary: Scene as Sparse 3D Gaussians}
\label{sec:method_preliminary}
Inspired by generalizable 3D reconstruction methods~\cite{pixelnerf, pixelsplat} which predict Gaussian parameters in a feed-forward manner, we utilize sparse feed-forward Gaussians to represent the scenes, akin to the Gaussians employed in previous work~\cite{gausstr, gaussianflowocc}. 
Each Gaussian $\mathbf{G}$ is parameterized by its 3D mean \mbox{$\mathbf{m}_{ego} \in \mathbb{R}^3$} in ego space, a 3D covariance matrix $\mathbf{\Sigma}$ composed of a rotation quaternion \mbox{$\mathbf{r} \in \mathbb{R}^4$} and a scaling factor \mbox{$\mathbf{s} \in \mathbb{R}^3$}, an opacity \mbox{$\alpha \in [0, 1]$}, and a high-dimensional feature vector \mbox{$\mathbf{f} \in \mathbb{R}^{C_{f}}$}.
For initialization, we first assign a 2D pixel location \mbox{$\mathbf{m}_{c}^{2D}=(u, v)$} and retrieve its corresponding depth value $d$ from a monocular estimated depth map $\mathbf{D}$, which can be optionally refined by LiDAR projected depth. These are combined into a 3D mean \mbox{$\mathbf{m}_{c}^{3D}=(u, v, d)$} in the image frame, and then transformed to the ego frame using the camera intrinsics $K$ and extrinsics $E_{e2c}$:
\begin{align}
    \mathbf{m}_{ego} = E_{e2c}^{-1}K^{-1}\mathbf{m}_{c}^{3D}.
    \label{eq: mean_transform}
\end{align}
The rotation quaternion $\mathbf{r}$ is initialized to be orthogonal to the camera view, while the scaling factor $\mathbf{s}$ is adjusted to be proportional to $d$ based on perspective principles. The opacity and feature vector are left uninitialized. Thus, 3D Gaussian primitive can be parameterized and initialized as:
\begin{align}
    \mathbf{G} = (\mathbf{m}_{ego}, \mathbf{s}, \mathbf{r}, \alpha, \mathbf{f}).
    \label{eq: gaussian_parameterization}
\end{align}

\begin{figure*}[htp] 
    \centering
    \includegraphics[height=4.5cm]{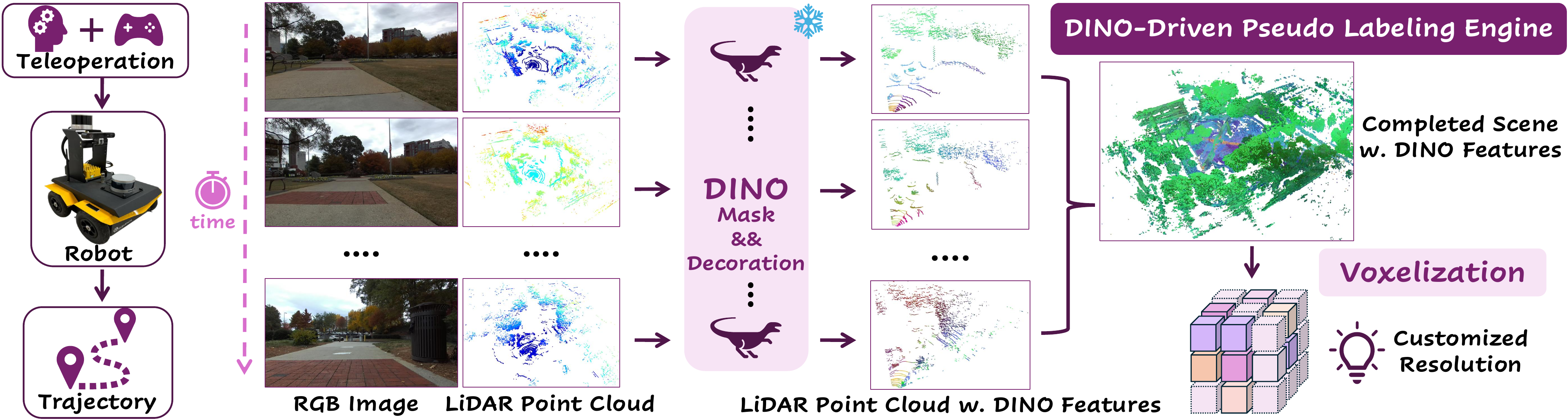}
    \vspace{-5pt}
    \caption{\textbf{Overview of DINO-Driven Pseudo Labeling Engine.} We teleoperate our UGV through urban scenarios to collect paired image and point cloud sequences along with trajectories from onboard camera and LiDAR. LiDAR points are then projected to image and decorated with pixel-wise DINO features. These points are aggregated and voxelized at a customized resolution to be 3D pseudo labels.}
    \vspace{-8pt}
    \label{fig:pseud_label}
\end{figure*}

\subsection{Multi-Modal Gaussian Transformer}
\label{sec:method_transformer}
Multi-modal transformer~\cite{transformer} has achieved great success in learning unified BEV or voxel representations from multiple sensors (e.g., LiDAR, camera) for 3D object detection~\cite{bevfusion, bevfusion_peking, unitr, futr3d, uvtr, transfusion, deepinteraction}. Motivated by this success, we propose a novel and generalizable multi-modal transformer for Gaussian-based 3D scene understanding. The learning objectives of our framework consist of a set of learnable Gaussian queries \mbox{$\mathbf{Q} \in \mathbb{R}^{N_g \times C_{g}}$} and their 3D mean positions \mbox{$\mathbf{M}_{c}^{3D} \in \mathbb{R}^{N_g \times 3}$} in the image frame, where $N_g$ and $C_{g}$ denote the number of Gaussians and the embedding dimension, respectively. We first extract feature maps from camera, LiDAR and radar data separately. For camera data, multi-scale feature maps $\mathbf{F}_{c}$ are extracted via a pre-trained DINO~\cite{dinov2, dinov3} backbone followed by a feature pyramid network (FPN)~\cite{fpn}. For LiDAR data, multi-scale BEV feature maps $\mathbf{F}_{l}$ are extracted through a 3D backbone SECOND~\cite{second} and FPN~\cite{fpn}. For radar data, its BEV feature map $\mathbf{F}_{r}$ is generated from a 3D backbone PointPillars~\cite{pointpillars}. We then transform $\mathbf{M}_{c}^{3D}$ into the LiDAR frame using the camera intrinsics $K$ and extrinsics $E_{l2c}$ to obtain 3D positions of Gaussian queries on LiDAR and radar feature maps:
\begin{align}
    \mathbf{M}_{l}^{3D} = E_{l2c}^{-1}K^{-1}\mathbf{M}_{c}^{3D}.
    \label{eq: pixel2lidar}
\end{align}
Afterwards, through a sequence of transformer decoder layers, we gradually aggregate features from different sensor modalities via deformable attention~\cite{deformable_detr}:
\begin{align}
    \mathbf{Q}_{i} &= \text{DeAttn}(\mathbf{F}_{i}, \mathbf{Q}, \mathbf{M}_{i}^{3D}), \hspace{3pt} i \in \{c, l, r\}.
    \label{eq: deform}
\end{align}
The sampled features are then concatenated and fused via a multi-layer perceptron (MLP) to update the queries:
\begin{align}
    \mathbf{Q} = \mathbf{Q} + \text{MLP}(\mathbf{Q}_{c} \oplus \mathbf{Q}_{l} \oplus \mathbf{Q}_{r}).
    \label{eq: fusion}
\end{align}
After cross-attention to multi-modal features, we further update Gaussian queries through self-attention layers~\cite{transformer} to interact Gaussian features across the scene:
\begin{align}
    \mathbf{Q} = \text{SelfAttn}(\mathbf{Q}, \text{PE}(\mathbf{M}_{c}^{3D})),
    \label{eq: self}
\end{align}
where $\text{PE}(\cdot)$ represents positional encoding. Finally, a Gaussian-wise nonlinear mapping is applied to further enhance Gaussian queries via a feed-forward network (FFN):
\begin{align}
    \mathbf{Q} = \text{FFN}(\mathbf{Q}).
    \label{eq: ffn}
\end{align}
The query positions $\mathbf{M}_{c}^{3D}$ are subsequently updated via a MLP by \mbox{$\mathbf{M}_{c}^{3D} = \mathbf{M}_{c}^{3D} + \text{MLP}(\mathbf{Q})$} within the decoder. To obtain other Gaussian properties, a set of MLPs are employed to regress these attributes based on learned queries:
\begin{align}
    \{\Delta s_{i}, \alpha_{i}, \mathbf{f}_{i}\} = \text{MLP}(\mathbf{Q}), \hspace{3pt} i \in \{1, \dots, N_g\}.
    \label{eq: head}
\end{align}
The final predicted Gaussians can be passed to different heads to enable downstream perception and planning tasks.

\subsection{Shelf-Supervised Learning for Gaussians}
\label{sec:method_shelf}

Instead of only aligning 2D Gaussian renderings with VFM features, we propose to jointly align 2D image-level and 3D scene-level Gaussian outputs with VFM features to narrow the performance gap to fully supervised methods.

\subsubsection{2D Image-Level Alignment with VFMs}
Leveraging the high-fidelity rendering capability of 3D Gaussians, we first splat the Gaussians onto source views and then align the renders with estimated depth maps and VFM features~\cite{gausstr}. To improve rendering efficiency, principal component analysis (PCA)~\cite{pca} is used to compress both the high-dimensional Gaussian features $\mathbf{f}$ into $\mathbf{\bar{f}}$, and the target feature maps $\mathbf{F}^{tgt}$ into $\mathbf{\bar{F}}^{tgt}$. The rendered depth map $\mathbf{\bar{D}}$ and feature map $\mathbf{\bar{F}}$ are then computed by alpha blending:
\begin{align}
    \mathbf{\bar{F}} = \sum_{i=1}^{N} \mathbf{\bar{f}}_i \alpha_i \prod_{j=1}^{i-1} (1 - \alpha_j),\\
    \mathbf{\bar{D}} = \sum_{i=1}^{N} d_i \alpha_i \prod_{j=1}^{i-1} (1 - \alpha_j),
    \label{eq: alpha}
\end{align}
where $d_i$ denotes the distance of each Gaussian to the camera along the viewing direction.
Afterwards, a cosine similarity-based feature loss $\mathcal{L}_{feat}^{2D}$, a $\mathrm{L}1$ depth loss $\mathcal{L}_{depth}^{2D}$, and a Scale-Invariant Logarithmic (SILog) depth loss~\cite{silogloss} $\mathcal{L}_{SILog}^{2D}$ are combined for 2D image-level supervision.

\subsubsection{3D Scene-Level Alignment with VFMs}
Current Gaussian-based approaches are supervised by either 2D feature maps or 3D closed-set labels, which present a trade-off and cannot achieve high efficiency and superior performance simultaneously. To alleviate this, we propose a \textit{DINO-Driven Pseudo Labeling Engine} to automatically generate high-fidelity 3D occupancy pseudo labels, and a \textit{CUDA-Accelerated Gaussian-to-Voxel Splatting} module to enable highly efficient Gaussian-to-voxel splatting, capable of supporting both forward and backward propagation of 3D Gaussians with high-dimensional features.

\noindent \textbf{DINO-Driven Pseudo Labeling Engine.} As shown in~\cref{fig:pseud_label}, the pseudo labeling engine begins by collecting paired RGB image and LiDAR point cloud sequences using onboard sensors (e.g.: LiDAR, camera) on a driving vehicle or mobile robot. DINO~\cite{dinov2, dinov3} is then employed to \emph{decorate} the raw LiDAR point clouds via point-to-image projection, i.e., points are projected onto 2D image plane to retrieve their corresponding VFM features. Notably, we apply a visibility mask to filter out points outside of the camera's field of view. Afterwards, LiDAR points of key frames are then accumulated for scene reconstruction. Lastly, the completed scene with DINO features is voxelized into pseudo occupancy labels at customized resolution. Our proposed pipeline can be inserted into any occupancy framework for automatic and high-quality 3D VFM feature labeling.

\begin{figure}[t] 
    \centering
    \includegraphics[width=\columnwidth]{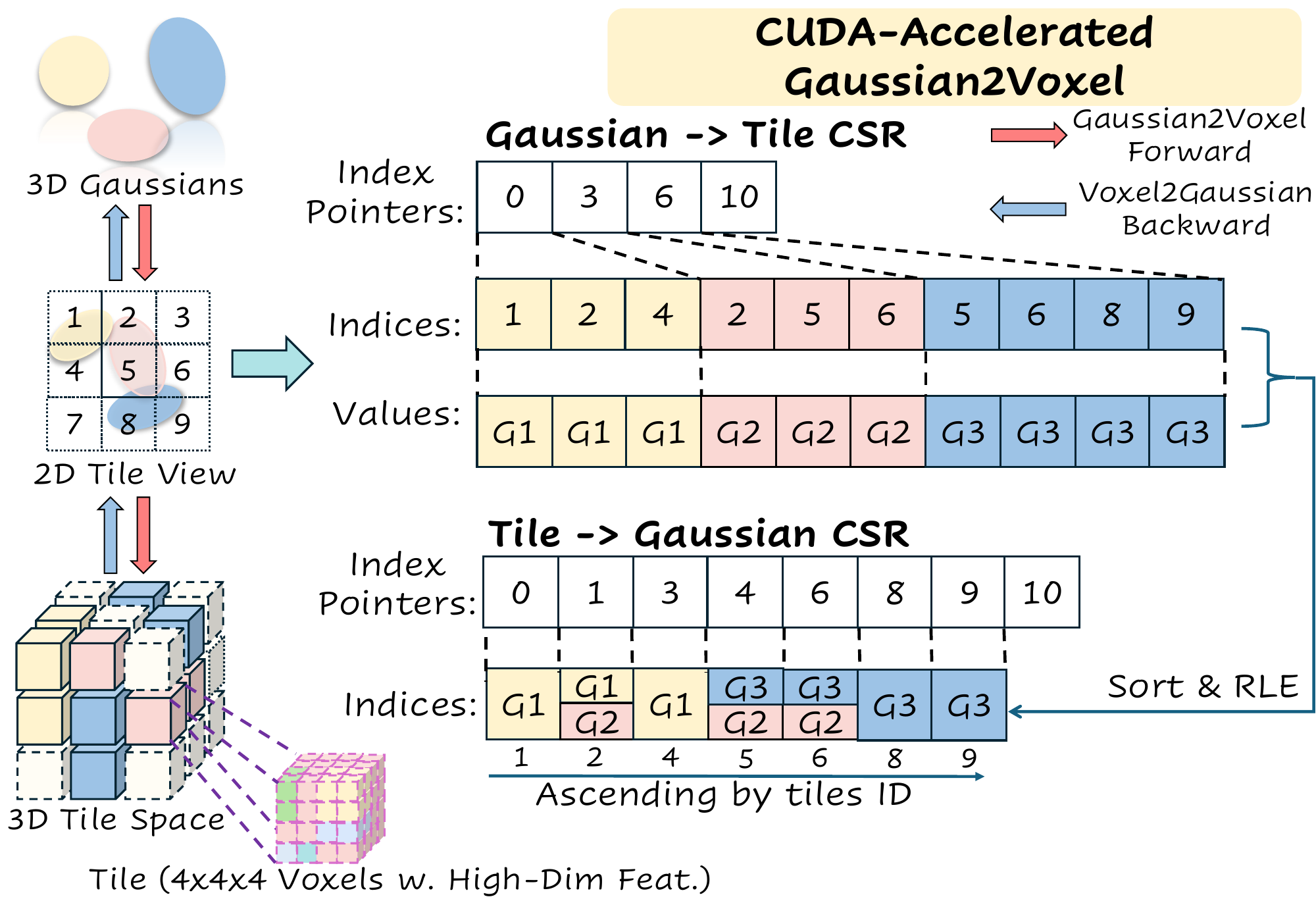}
    \vspace{-15pt}
    \caption{\textbf{Dual-CSR Structure for CUDA-Accelerated Gaussian2Voxel}. 
    \textbf{Gaussian$\rightarrow$Tile CSR:} index pointers store tile offsets per Gaussian, indices record tile IDs, and values store Gaussian IDs. \textbf{Tile$\rightarrow$Gaussian CSR:} index pointers store Gaussian offsets per tile, and indices record Gaussian IDs obtained by sorting and run-length encoding (RLE) tile-Gaussian pairs.}
    \vspace{-8pt}
    \label{fig:cuda_gaussian2voxel}
\end{figure}

\noindent \textbf{CUDA-Accelerated Gaussian-to-Voxel Splatting.}
An efficient Gaussian-to-Voxel splatting module can significantly reduce overall training time. However, PyTorch-based implementations in GaussTR~\cite{gausstr} remain computationally expensive, and the CUDA-based implementation in GaussianFormer~\cite{gaussianformer} is primarily designed for probability computation with a small set of semantic categories and only supports operations for Gaussians centered within a fixed voxel-space range. As a result, Gaussians that are located outside this range while still overlapping with target voxels are ignored, leading to inaccurate G2V conversion.
In contrast, our approach is able to accurately and efficiently account for contributions from all Gaussians, regardless of their spatial locations. To this, we implement an efficient computation strategy based on a dual Compressed Sparse Row (CSR)~\cite{csr} structure for mapping between Gaussians and tiles, where each tile is defined as a block of $4\times4\times4$ voxels. 
As illustrated in ~\cref{fig:cuda_gaussian2voxel}, for the forward, based on Tile-to-Gaussian CSR, we assign one tile to each CUDA block for parallel voxel accumulation, which optimizes \text{L1} cache usage and reduces redundant access. For the backward, based on Gaussian-to-Tile CSR, we assign one Gaussian per block to propagate gradients without atomic operations, thereby improving computational efficiency.

\noindent
\textbf{3D Shelf-Supervision.} For 3D supervision, based on learned Gaussian properties and embeddings, ShelfGaussian first predicts occupancy densities $\boldsymbol{\rho} \in \mathbb{R}^{N_v \times 1}$ and corresponding VFM features \mbox{$\mathbf{O} \in \mathbb{R}^{N_v \times C_v}$}, where $N_v$ and $C_v$ denote the number of voxels and feature dimension. An MLP is then adopted to regress binary occupancy logits via \mbox{$\mathbf{o}=\text{MLP}(\boldsymbol{\rho})$}, and a predicted binary occupancy mask $\mathbf{\hat{o}}^{pred} \in \mathbb{R}^{N_v \times 1}$ is obtained by \mbox{$\mathbf{\hat{o}}^{pred} = \mathbf{o} > \tau$}, where $\tau$ is an empirical threshold. By multiplying $\mathbf{\hat{o}}^{pred}$ with pseudo binary occupancy label $\mathbf{\hat{o}}^{pseudo}$ and camera visibility mask $\mathbf{\hat{o}}^{vis}$, we derive the final feature supervision mask $\mathbf{\hat{o}}$. The final predicted occupancy features are formed by \mbox{$\mathbf{O}^{pred} = \mathbf{\hat{o}} \times \mathbf{O}$}. A binary cross entropy loss $\mathcal{L}_{bce}^{3D}$ between $\mathbf{o}$ and $\mathbf{\hat{o}}^{p}$, and a cosine similarity-based feature loss $\mathcal{L}_{feat}^{3D}$ between $\mathbf{O}^{pred}$ and pseudo feature label $\mathbf{O}^{pseudo}$ are applied for 3D supervision. Finally, by combining 2D and 3D losses, we train the model with the overall loss as follows:
\begin{align}
    \mathcal{L} = \mathcal{L}_{feat}^{2D} + \mathcal{L}_{depth}^{2D} + \mathcal{L}_{SILog}^{2D} + \mathcal{L}_{bce}^{3D} + \mathcal{L}_{feat}^{3D}.
    \label{eq: overall_loss}
\end{align}

\subsection{Multi-Task Evaluation}
\label{sec:method_evaluation}
\subsubsection{Zero-Shot 3D Scene Understanding}
We evaluate the learned VFM-aligned Gaussian representations on zero-shot semantic occupancy prediction and BEV map segmentation. During inference, we first leverage pre-trained models from Talk2DINO~\cite{talk2dino} or \text{DINO.txt}~\cite{dinotxt} to map the CLIP~\cite{clip} text embeddings of open-set queries into the DINO feature space, denoted as \mbox{$\mathbf{f}_{txt} \in \mathbb{R}^{N_c \times C_f}$}, where $N_c$ is the number of semantic categories. Then we measure the cosine similarity between Gaussian features $\mathbf{f}$ and text embeddings $\mathbf{f}_{txt}$ to calculate the semantic logits for the Gaussian by \mbox{$\mathbf{s} = \text{Softmax}(\mathbf{f} \cdot \mathbf{f}_{txt}^{\text{T}})$}. Finally, we voxelize the Gaussians into a target resolution for zero-shot semantic occupancy prediction and BEV segmentation via Gaussian-to-voxel and Gaussian-to-BEV splatting separately.

\subsubsection{Gaussian-based Trajectory Planning}
Most existing methods adopt BEV~\cite{uniad, stp3, vad, bevplanner} or voxel~\cite{occnet, occworld, occvla} representations for end-to-end trajectory planning. Recent frameworks such as GaussianFusion~\cite{gaussianfusion} and GaussianAD~\cite{gaussianad} adopt a Gaussian representation similar to that in GaussianFormer~\cite{gaussianformer}, trained with auxiliary perception loss based on closed-set human-annotated labels. In contrast, we propose a simple yet effective \emph{Gaussian-Planner} framework to directly apply our shelf-supervised open-vocabulary Gaussians to open-loop trajectory planning. To illustrate, we first initialize an ego query \mbox{$\mathbf{Q}_{ego} \in \mathbb{R}^{N_{ego} \times C_{ego}}$} as a learnable embedding, where $N_{ego}$ and $C_{ego}$ denote the number of ego queries and query dimension, respectively. We then perform cross attention~\cite{transformer} between ego query and learned Gaussian embeddings $\mathbf{Q}$ to update the ego query. Finally, the trajectory is predicted by decoding the updated ego query with a MLP. The ego status (e.g., velocity, acceleration) can be concatenated to the ego query optionally for further enhancing vehicle self-awareness and motion feasibility. The prediction of future trajectory \mbox{$\mathbf{X} \in \mathbb{R}^{N_{step} \times 2}$} can be formulated as follows:
\begin{align}
    \mathbf{X} = \text{MLP}(\text{CrossAttn}(\mathbf{Q}_{ego}, \mathbf{Q})).
    \label{eq: planning}
\end{align}
An $\mathrm{L}_{1}$ loss between predicted and ground-truth trajectories is adopted for supervision.

\section{Experiment}
\label{sec:experiment}

\definecolor{nbarrier}{RGB}{112, 128, 144}
\definecolor{nbicycle}{RGB}{220, 20, 60}
\definecolor{nbus}{RGB}{255, 127, 80}
\definecolor{ncar}{RGB}{255, 158, 0}
\definecolor{nconstruct}{RGB}{233, 150, 70}
\definecolor{nmotor}{RGB}{255, 61, 99}
\definecolor{npedestrian}{RGB}{0, 0, 230}
\definecolor{ntraffic}{RGB}{47, 79, 79}
\definecolor{ntrailer}{RGB}{255, 140, 0}
\definecolor{ntruck}{RGB}{255, 99, 71}
\definecolor{ndriveable}{RGB}{0, 207, 191}
\definecolor{nother}{RGB}{175, 0, 75}
\definecolor{nsidewalk}{RGB}{75, 0, 75}
\definecolor{nterrain}{RGB}{112, 180, 60}
\definecolor{nmanmade}{RGB}{222, 184, 135}
\definecolor{nvegetation}{RGB}{0, 175, 0}
\definecolor{nothers}{RGB}{0, 0, 0}

\begin{table*}[!h]
  \centering
  \resizebox{\linewidth}{!}{
    \begin{tabular}{c|c|>{\columncolor{black!10}}c|>{\columncolor{black!10}}c|ccccccccccccccccc}
      \toprule
      Method & Mod. & IoU  & mIoU & 
      \rotatebox{90}{others} &
      \rotatebox{90}{barrier} & 
      \rotatebox{90}{bicycle} & 
      \rotatebox{90}{bus} & 
      \rotatebox{90}{car} & 
      \rotatebox{90}{const. veh.} & 
      \rotatebox{90}{motorcycle} & 
      \rotatebox{90}{pedestrian} & 
      \rotatebox{90}{traffic cone} & 
      \rotatebox{90}{trailer} & 
      \rotatebox{90}{truck} & 
      \rotatebox{90}{drive. surf.} & 
      \rotatebox{90}{other flat} & 
      \rotatebox{90}{sidewalk} & 
      \rotatebox{90}{terrain} & 
      \rotatebox{90}{manmade} & 
      \rotatebox{90}{vegetation} \\
      &  & & &
    \tikz \draw[fill=black,draw=black] (0,0) rectangle (0.2,0.2); &      
      \tikz \draw[fill=nbarrier,draw=nbarrier] (0,0) rectangle (0.2,0.2); & 
      \tikz \draw[fill=nbicycle,draw=nbicycle] (0,0) rectangle (0.2,0.2); & 
      \tikz \draw[fill=nbus,draw=nbus] (0,0) rectangle (0.2,0.2); & 
      \tikz \draw[fill=ncar,draw=ncar] (0,0) rectangle (0.2,0.2); & 
      \tikz \draw[fill=nconstruct,draw=nconstruct] (0,0) rectangle (0.2,0.2); & 
      \tikz \draw[fill=nmotor,draw=nmotor] (0,0) rectangle (0.2,0.2); & 
      \tikz \draw[fill=npedestrian,draw=npedestrian] (0,0) rectangle (0.2,0.2); & 
      \tikz \draw[fill=ntraffic,draw=ntraffic] (0,0) rectangle (0.2,0.2); & 
      \tikz \draw[fill=ntrailer,draw=ntrailer] (0,0) rectangle (0.2,0.2); & 
      \tikz \draw[fill=ntruck,draw=ntruck] (0,0) rectangle (0.2,0.2); & 
      \tikz \draw[fill=ndriveable,draw=ndriveable] (0,0) rectangle (0.2,0.2); & 
      \tikz \draw[fill=nother,draw=nother] (0,0) rectangle (0.2,0.2); & 
      \tikz \draw[fill=nsidewalk,draw=nsidewalk] (0,0) rectangle (0.2,0.2); & 
      \tikz \draw[fill=nterrain,draw=nterrain] (0,0) rectangle (0.2,0.2); & 
      \tikz \draw[fill=nmanmade,draw=nmanmade] (0,0) rectangle (0.2,0.2); & 
      \tikz \draw[fill=nvegetation,draw=nvegetation] (0,0) rectangle (0.2,0.2); \\
      \midrule
      \multicolumn{21}{l}{\textit{Supervised by 2D Image}}  \\
		\midrule
        SimpleOcc~\cite{simpleocc} & C & - & 7.05& 0.00& 0.67& 1.18& 3.21& 7.63& 1.02& 0.26& 1.80& 0.26& 1.07& 2.81& 40.44& 0.00& 18.30& 17.01& 13.42& 10.84\\
		SelfOcc~\cite{selfocc} & C & 45.01  & 9.30 & 0.00 & 0.15 & 0.66 & 5.46 & 12.54 & 0.00 & 0.80 & 2.10 & 0.00 & 0.00 & 8.25 & 55.49 & 0.00 & 26.30 & 26.54 & 14.22 & 5.60 \\
		OccNeRF~\cite{occnerf} & C & 22.81 & 9.53 & 0.00 & 0.83 & 0.82 & 5.13 & 12.49 & 3.50 & 0.23 & 3.10 & 1.84 & 0.52 & 3.90 & 52.62 & 0.00 & 20.81 & 24.75 & 18.45 & 13.19 \\
        DistillNeRF~\cite{distillnerf} & C & 29.11 & 8.93 & 0.03&  1.35 & 2.08 & 10.21 & 10.09 & 2.56 & 1.98 & 5.54 & 4.62 & 1.43 & 7.90 & 43.02 & 0.00 & 16.86 & 15.02 & 14.06 & 15.06 \\
        GaussianOcc~\cite{gaussianocc} & C & - & 9.94& 0.00 &1.79& 5.82& 14.58& 13.55& 1.30& 2.82& 7.95& \underline{9.76}& 0.56& 9.61& 44.59& 0.00&20.10& 17.58& 8.61& 10.29\\
        LangOcc~\cite{langocc} & C & 51.76 & 11.84& 0.00& 3.10& 9.00& 6.30& 14.20& 0.40& 10.80& 6.20& 9.00& 3.80& 10.70& 43.70& \textbf{2.23}& 9.50& 26.40& 19.60& 26.40\\
        GaussTR~\cite{gausstr} & C & 44.54& 12.27& 0.00& 6.50& 8.54& 21.77& 24.27& 6.26& 15.48& 7.94& 1.86& 6.10& 17.16& 36.98& 0.00& 17.21& 7.16& 21.18& 9.99 \\
      \midrule
	   \multicolumn{21}{l}{\textit{Supervised by 3D Pseudo Label or Temporal Information}}  \\
	   \midrule
       VEON-B~\cite{veon} & C & - & 12.38 & 0.50& 4.80& 2.70& 14.70& 10.90& 11.00& 3.80& 4.70& 4.00& 5.30& 9.60& 46.50& \underline{0.70}& 21.10& 22.10& 24.80& 23.70\\
       GaussianFlowOcc~\cite{gaussianflowocc} & C & 46.91& 17.08& 0.00& \underline{6.75} & 9.68& 18.98& 17.15& 4.19& 11.78& 9.27& \textbf{10.30}& 1.83& 12.33& \underline{61.03}& 0.00& \underline{31.17}& \textbf{34.78}& 14.66& 12.40 \\
       LOcc~\cite{locc} & C & - & 18.62& 0.00& \textbf{14.21} & 0.90& 27.21& \textbf{32.61} & 3.09& 2.06& 8.67& 1.74& 1.96& 21.45& \textbf{71.22}& 0.00 &\textbf{38.18} & \underline{33.32} &30.68& 29.32 \\
       \midrule
      \multirow{5}{*}{ShelfGaussian} & C & 63.25 & 19.07 & 0.36 & 2.96 & 10.84 & 30.20 & 30.20 & 11.34 & 18.93 & 14.49 & 6.54 & 9.47 & 17.77 & 60.02 & 0.14 & 29.66 & 21.62 & 29.63 & 30.08 \\
      ~ & L & 66.10 & 19.34 & 0.23 & 3.70 & 4.17 & 31.25 & 29.87 & 15.12 & 8.65 & \textbf{24.34} & 2.87 & \textbf{14.62} & 22.06 & 53.34 & 0.06 & 23.97 & 18.63 & \textbf{36.02} & \textbf{39.96}\\
      ~ & C+R & 62.84 & 19.42 & 0.51 & 2.77 & 11.72 & 31.15 & 30.46 & 12.40 & 19.56 & 15.35 & 6.60 & 11.19 & 19.28 & 58.89 & 0.15 & 28.17 & 21.36 & 29.58 & 31.01\\
      ~ & L+C & \underline{69.24} & \underline{21.52} & \textbf{0.60} & 3.08 & \underline{11.80} & \textbf{36.36} & 31.40 & \textbf{14.71} & \underline{21.04} & 18.57 & 6.62 & \underline{14.44} & \underline{22.85} & 58.43 & 0.15 & 28.45 & 22.88 & 35.34 & 39.11\\
      ~ & L+C+R & \textbf{69.45} & \textbf{21.78} & \underline{0.58} & 3.30 & \textbf{12.16} & \underline{34.20} & \underline{31.62} & \underline{14.59} & \textbf{21.84} & \underline{20.88} & 7.06 & 14.24 & \textbf{22.89} & 59.91 & 0.14 & 29.14 & 22.77 & \underline{35.35} & \underline{39.64}\\

      \bottomrule
    \end{tabular}
  }
  \vspace{-5pt}
  \caption{\textbf{Zero-shot semantic occupancy prediction performance of ShelfGaussian on Occ3D-nuScenes~\cite{occ3d} dataset.} C, L and R denote camera, LiDAR and radar for simplicity. Mod. denotes the modality input during \textit{inference}. The best is in \textbf{bolded} and the second best is \underline{underlined}. Ours achieves the state-of-the-art performance with least number of scene queries, and works for any modality as input.}
  \vspace{-10pt}
  \label{tab:occ3d}
\end{table*}
\begin{table}[!h]
\centering
\resizebox{\linewidth}{!}{
\begin{tabular}{c|c|>{\columncolor{black!10}}c|>{\columncolor{black!10}}c|ccc}

\midrule
\multirow{2}{*}{Method} & 
\multirow{2}{*}{Mod.} &
~&
~ &
\multicolumn{3}{c}{IoU}
\\
~ & ~ & \multirow{-2}{*}{GT}  & \multirow{-2}{*}{Sup.}  & Vehicle & Pedestrian & Drivable Area \\
\midrule
LSS~\cite{lss} & C & \checkmark & FS & - & 15.0 & 72.9 \\
FIERY~\cite{fiery} & C & \checkmark & FS & 35.8 & 17.2 & -\\
BEVFormer~\cite{bevformer} & C & \checkmark & FS & 35.8 & - & 80.1\\
PointBeV~\cite{pointbev} & C & \checkmark & FS & 38.7 & 18.5 & 72.5\\
GaussianBEV~\cite{gaussianbev} & C & \checkmark & FS & \textbf{41.6} & \textbf{21.2} & \textbf{82.6} \\

\midrule
ShelfGaussian  & C & \ding{55} & ZS & \textbf{21.1} & \textbf{6.2} & \textbf{38.4}\\

\hline

\end{tabular}
}
\vspace{-5pt}
\caption{\textbf{Performance of ShelfGaussian on zero-shot BEV map segmentation on nuScenes~\cite{nuscenes} dataset.} GT: ground-truth. Sup.: supervision. FS: fully-supervised. ZS: zero-shot. Results of other methods are reported in~\cite{gaussianbev}.}
\vspace{-5pt}
\label{tab:bev_seg}
\end{table}

\begin{table}[!h]
\centering
\resizebox{\linewidth}{!}{
\begin{tabular}{c|ccc>{\columncolor{black!10}}c|ccc>{\columncolor{black!10}}c}

\midrule
\multirow{2}{*}{Method} & 
\multicolumn{4}{c|}{L2 (m) $\downarrow$} &
\multicolumn{4}{c}{CR (\%) $\downarrow$} \\
   & 
 1s & 2s & 3s & Avg. &
1s & 2s & 3s & Avg. 
\\
\midrule
 ST-P3~\cite{stp3}   & 1.59 & 2.64 & 3.73 & 2.65
   & 0.69 & 3.62 & 8.39 & 4.23  \\

 UniAD*~\cite{uniad}    & 0.20& 0.42& 0.75& 0.46& 0.02& 0.25& 0.84& 0.37\\

 VAD-Base*~\cite{vad} & 0.17& 0.34& 0.60& 0.37& 0.04& 0.27& \textbf{0.67}& 0.33 \\


OccWorld~\cite{occworld} & 0.52 & 1.27 & 2.41 & 1.40 & 0.12 & 0.40 & 2.08 & 0.87\\

OccNet~\cite{occnet}  & 1.29 & 2.13 & 2.99 & 2.14 & 0.21 & 0.59 & 1.37 & 0.72\\

GaussianAD(5f)~\cite{gaussianad}  & 0.40 & 0.64 & 0.88 & 0.64 & 0.09 & 0.38 & 0.81 & 0.42\\
 
 BEV-Planner(4f)*~\cite{bevplanner} &
\textbf{0.16} & \textbf{0.32} & \textbf{0.57} & \textbf{0.35} &
0.04 & 0.29 & 0.94 & 0.42 \\

\midrule
Gaussian-Planner* & \textbf{0.16} & \textbf{0.32} & 0.58 & \textbf{0.35} & \textbf{0.00} & \textbf{0.18} & 0.78 & \textbf{0.32} \\

\hline

\end{tabular}
}
\vspace{-5pt}
\caption{\textbf{Performance comparison of open-loop trajectory planning methods on nuScenes~\cite{nuscenes} dataset.} (xf) and * denote the number of past frames and ego trajectory in usage, respectively. $\mathrm{L}$2 error and CR are computed the same way as ST-P3~\cite{stp3}.}
\vspace{-10pt}
\label{tab:open_loop_planning}
\end{table}

\subsection{Datasets and Metrics}
\textbf{Datasets.}
\textbf{NuScenes}~\cite{nuscenes} dataset provides 1000 sequences of driving scenes collected with 6 surrounding cameras, 1 LiDAR, 5 radars and 1 IMU. Occ3D~\cite{occ3d} offers semantic occupancy annotation across 18 classes for nuScenes dataset within the range of $[-40\rm{m}, 40\rm{m}] \times [-40\rm{m}, 40\rm{m}] \times [-1\rm{m}, 5.4\rm{m}]$ with a voxel resolution of $0.4\mathrm{m}$. To evaluate the in-the-wild performance, we built a \textbf{Custom Dataset} by collecting paired monocular RGB images and LiDAR point clouds across diverse urban scenes using a Clearpath Jackal robot. We use a ZED 2i stereo camera to collect images and a combination of Velodyne VLP16 and Hesai XT16 LiDARs for point clouds. Our custom dataset consists of $2638$ training and $294$ testing frames, collected in \textit{garden}, \textit{street}, \textit{park} and \textit{grassland} scenarios. Pseudo occupancy labels are generated at 2Hz in a range of $[0\rm{m}, 20\rm{m}] \times [-10\rm{m}, 10\rm{m}] \times [-1\rm{m}, 4\rm{m}]$ with a voxel resolution of $0.2\mathrm{m}$.

\noindent \textbf{Metrics.}
Intersection-over-Union (IoU) and mean IoU (mIoU) across all classes are used for evaluation of our model on zero-shot semantic occupancy prediction and BEV map segmentation. $\mathrm{L}2$ error and collision rate (CR) are reported to assess the trajectory planning performance. 
$\mathrm{L}2$ error is defined as the mean Euclidean distance between predicted and ground-truth trajectories while CR is the percentage of the planned trajectory intersecting an obstacle.

\begin{figure*}[htp] 
    \centering
    \includegraphics[height=7cm]{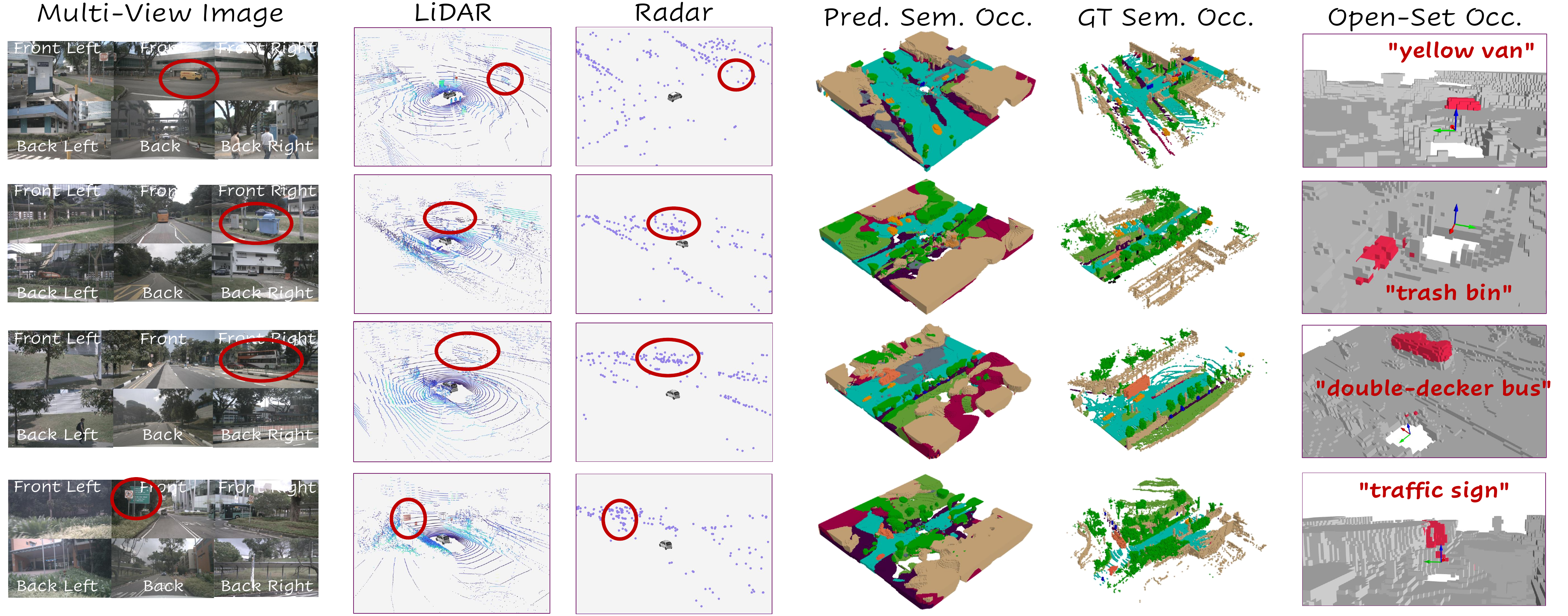}
    \vspace{-15pt}
    \caption{\textbf{Qualitative results of ShelfGaussian on nuScenes dataset.} The figure demonstrates the predicted semantic occupancy queried by semantic classes in~\cref{tab:occ3d}, ground-truth labels from Occ3D~\cite{occ3d} and occupancy of open-set queries from ShelfGaussian-LCR model. Best viewed on screen and color bar is given in~\cref{tab:occ3d}.}
    \vspace{-5pt}
    \label{fig:qualitative_nuscenes}
\end{figure*}

\definecolor{nbarrier}{RGB}{112, 128, 144}
\definecolor{nbicycle}{RGB}{220, 20, 60}
\definecolor{nbus}{RGB}{255, 127, 80}
\definecolor{ncar}{RGB}{255, 158, 0}
\definecolor{nconstruct}{RGB}{233, 150, 70}
\definecolor{nmotor}{RGB}{255, 61, 99}
\definecolor{npedestrian}{RGB}{0, 0, 230}
\definecolor{ntraffic}{RGB}{47, 79, 79}
\definecolor{ntrailer}{RGB}{255, 140, 0}
\definecolor{ntruck}{RGB}{255, 99, 71}
\definecolor{ndriveable}{RGB}{0, 207, 191}
\definecolor{nother}{RGB}{175, 0, 75}
\definecolor{nsidewalk}{RGB}{75, 0, 75}
\definecolor{nterrain}{RGB}{112, 180, 60}
\definecolor{nmanmade}{RGB}{222, 184, 135}
\definecolor{nvegetation}{RGB}{0, 175, 0}
\definecolor{nothers}{RGB}{0, 0, 0}

\begin{table*}[h]
  \centering
  \resizebox{\linewidth}{!}{
    \begin{tabular}{c|cc|c|c|>{\columncolor{cyan!7}}c>{\columncolor{cyan!7}}c>{\columncolor{orange!7}}c>{\columncolor{orange!7}}c>{\columncolor{cyan!7}}c>{\columncolor{cyan!7}}c>{\columncolor{orange!7}}c>{\columncolor{orange!7}}c>{\columncolor{cyan!7}}c>{\columncolor{cyan!7}}c>{\columncolor{orange!7}}c>{\columncolor{orange!7}}c>{\columncolor{cyan!7}}c>{\columncolor{orange!7}}c>{\columncolor{orange!7}}c>{\columncolor{orange!7}}c>{\columncolor{orange!7}}c>{\columncolor{orange!7}}c}
      \toprule
      Mod. & DINOv3 & DINOv2 & IoU  & mIoU & 
      \rotatebox{90}{others} &
      \rotatebox{90}{barrier} & 
      \rotatebox{90}{bicycle} & 
      \rotatebox{90}{bus} & 
      \rotatebox{90}{car} & 
      \rotatebox{90}{const. veh.} & 
      \rotatebox{90}{motorcycle} & 
      \rotatebox{90}{pedestrian} & 
      \rotatebox{90}{traffic cone} & 
      \rotatebox{90}{trailer} & 
      \rotatebox{90}{truck} & 
      \rotatebox{90}{drive. surf.} & 
      \rotatebox{90}{other flat} & 
      \rotatebox{90}{sidewalk} & 
      \rotatebox{90}{terrain} & 
      \rotatebox{90}{manmade} & 
      \rotatebox{90}{vegetation} \\
      &  & & & &
    \tikz \draw[fill=black,draw=black] (0,0) rectangle (0.2,0.2); &      
      \tikz \draw[fill=nbarrier,draw=nbarrier] (0,0) rectangle (0.2,0.2); & 
      \tikz \draw[fill=nbicycle,draw=nbicycle] (0,0) rectangle (0.2,0.2); & 
      \tikz \draw[fill=nbus,draw=nbus] (0,0) rectangle (0.2,0.2); & 
      \tikz \draw[fill=ncar,draw=ncar] (0,0) rectangle (0.2,0.2); & 
      \tikz \draw[fill=nconstruct,draw=nconstruct] (0,0) rectangle (0.2,0.2); & 
      \tikz \draw[fill=nmotor,draw=nmotor] (0,0) rectangle (0.2,0.2); & 
      \tikz \draw[fill=npedestrian,draw=npedestrian] (0,0) rectangle (0.2,0.2); & 
      \tikz \draw[fill=ntraffic,draw=ntraffic] (0,0) rectangle (0.2,0.2); & 
      \tikz \draw[fill=ntrailer,draw=ntrailer] (0,0) rectangle (0.2,0.2); & 
      \tikz \draw[fill=ntruck,draw=ntruck] (0,0) rectangle (0.2,0.2); & 
      \tikz \draw[fill=ndriveable,draw=ndriveable] (0,0) rectangle (0.2,0.2); & 
      \tikz \draw[fill=nother,draw=nother] (0,0) rectangle (0.2,0.2); & 
      \tikz \draw[fill=nsidewalk,draw=nsidewalk] (0,0) rectangle (0.2,0.2); & 
      \tikz \draw[fill=nterrain,draw=nterrain] (0,0) rectangle (0.2,0.2); & 
      \tikz \draw[fill=nmanmade,draw=nmanmade] (0,0) rectangle (0.2,0.2); & 
      \tikz \draw[fill=nvegetation,draw=nvegetation] (0,0) rectangle (0.2,0.2); \\
      \midrule
      \cellcolor{white} ~ & \checkmark & & 61.03 & 13.27 & 1.21 & 3.28  & 7.33  & 23.41 & 35.01 & 17.78               & 8.54     & 3.03     & 8.82       & 14.06  & 12.58 & 55.82            & 1.05     & 0.03   & 13.97  & 7.29  & 12.36\\
      \cellcolor{white} \multirow{-2}{*}{C} & & \checkmark & 63.25 & 19.07 & 0.36 & 2.96 & 10.84 & 30.20 & 30.20 & 11.34 & 18.93 & 14.49 & 6.54 & 9.47 & 17.77 & 60.02 & 0.14 & 29.66 & 21.62 & 29.63 & 30.08 \\
      \midrule
       \cellcolor{white} ~ & \checkmark & & 69.24 & 15.41 & 1.05 & 4.15  & 8.63  & 26.31 & 39.76 & 25.12               & 10.84     & 3.65     & 11.61       & 22.22  & 14.97 & 54.17            & 1.22     & 0.08   & 14.52  & 8.72  & 14.99\\
        \cellcolor{white} \multirow{-2}{*}{L+C} & & \checkmark & 69.24 & 21.52 & 0.60 & 3.08 & 11.80 & 36.36 & 31.40 & 14.71 & 21.04 & 18.57 & 6.62 & 14.44 & 22.85 & 58.43 & 0.15 & 28.45 & 22.88 & 35.34 & 39.11 \\
      \bottomrule
    \end{tabular}
  }
  \vspace{-5pt}
  \caption{\textbf{Ablation on alignment with different VFMs.} DINOv3~\cite{dinov3}: ViT-B/16, $432 \times 768$ image input. DINOv2~\cite{dinov2}: ViT-B/14, $378 \times 672$ image input. Better results with DINOv2 and DINOv3 are highlighted in \textcolor{orange}{orange} and \textcolor{cyan}{cyan} colors, respectively.}
  \vspace{-12pt}
  \label{tab:ablation_vfm}
\end{table*}
\begin{table}[!h]
\centering
\resizebox{\linewidth}{!}{
\begin{tabular}{c|ccc|cc}

\midrule
Mod. & 
2D Loss &
3D BCE Loss &
3D Feat. Loss &
IoU &
mIoU
\\
\midrule
\cellcolor{white}~& \checkmark & & & 47.26 & 12.39\\
\cellcolor{white}~ & \checkmark & \checkmark & & 59.92 & 14.49\\
\cellcolor{white}~ & & \checkmark & \checkmark & 58.64 & 18.41\\
\rowcolor{black!10}
\cellcolor{white} \multirow{-4}{*}{C}  & \checkmark & \checkmark & \checkmark & 63.25 & 19.07\\

\hline

\end{tabular}
}
\vspace{-7pt}
\caption{\textbf{Ablation on shelf-supervision loss.}}
\vspace{-7pt}
\label{tab:ablation_loss_tab}
\end{table}
\begin{table}[!h]
\centering
\resizebox{\linewidth}{!}{
\begin{tabular}{c|ccc|ccc}

\midrule
\multirow{2}{*}{Method} & 
\multicolumn{3}{c|}{18k Gauss. w. 1024D} &
\multicolumn{3}{c}{9k Gauss. w. 768D} \\
   & 
 FW(s)  & BW(s) & Mem.(GB) &
FW(s) & BW(s) & Mem.(GB) 
\\
\midrule
GaussTR~\cite{gausstr} & 14.1 & 596.7 & 22.3 & 12.4 & 446.4 & 14.8\\
GaussianFormer~\cite{gaussianformer} & 5.4 & 13.4 & 12.9 & 1.9 & 4.2 & 7.7\\
\midrule
\rowcolor{black!10}
ShelfGaussian & \textbf{0.5} & \textbf{2.7} & \textbf{4.9} & \textbf{0.2} & \textbf{1.0} & \textbf{3.7}\\

\hline

\end{tabular}
}
\vspace{-5pt}
\caption{\textbf{Performance comparison of G2V splatting module.} FW: forward time. BW: backward time. Mem.: memory. All the experiments are completed on one Nvidia L40S GPU.}
\vspace{-10pt}
\label{tab:g2v}
\end{table}

\begin{figure*}[htp] 
    \centering
    \includegraphics[height=7cm]{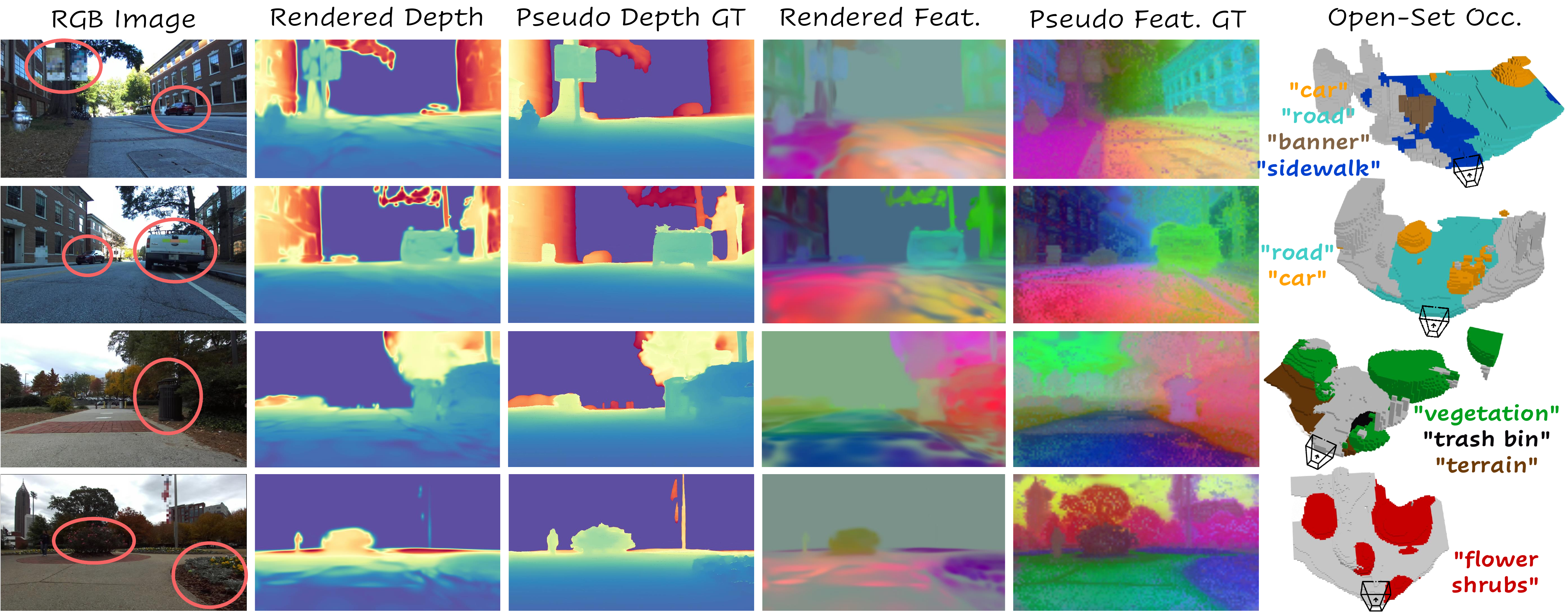}
    \vspace{-18pt}
    \caption{\textbf{Qualitative results of ShelfGaussian on custom dataset collected by a UGV.} The figure shows the rendered depth map, DINO feature map, and occupancy of novel categories from ShelfGaussian-CO model. Best viewed on screen and in color.}
    \vspace{-12pt}
    \label{fig:qualitative_custom}
\end{figure*}

\begin{figure}[t] 
    \centering
    \includegraphics[width=\columnwidth]{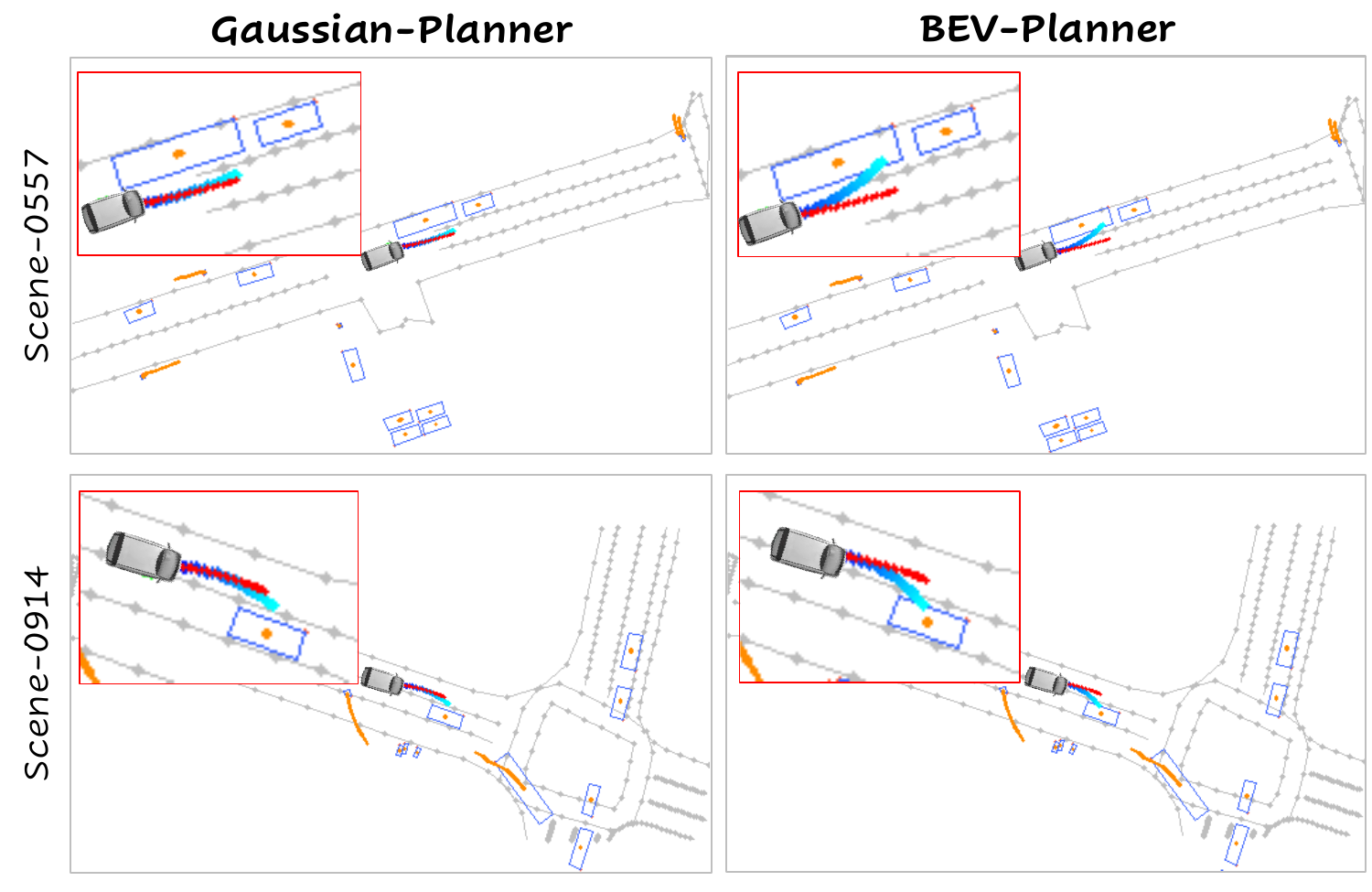}
    \vspace{-20pt}
    \caption{\textbf{Qualitative comparison of BEV-Planner~\cite{bevplanner} and Gaussian-Planner on nuScenes~\cite{nuscenes} dataset.} \textcolor{Red}{Red} and \textcolor{cyan}{cyan} lines denote the ground-truth and predicted trajectories separately.}
    \vspace{-10pt}
    \label{fig:planning}
\end{figure}

\subsection{Implementation Details}
We evaluate our method with DINOv2~\cite{dinov2} and DINOv3~\cite{dinov3} ViT-B models, and image resolutions are set as \mbox{$378 \times 672$} and \mbox{$432 \times 768$}, respectively. The DINO feature dimension is compressed to $128$ via PCA~\cite{pca}. Metric3Dv2~\cite{metric3dv2} is used for depth supervision. SECOND~\cite{second} and PointPillars~\cite{pointpillars} serve as the 3D backbone for LiDAR and radar, respectively. Talk2DINO~\cite{talk2dino} and DINO.txt~\cite{dinotxt} are utilized for vision-language alignment. The number of Gaussians per view is set as $1000$ in our main experiments. The Gaussian embedding dimension is set to $256$. We train our perception model for 24 epochs on 8 Nvidia L40S GPUs with a learning rate of $2 \times 10^{-4}$ and a batch size of 8. The planning model is further trained for 12 epochs on 8 Nvidia V100 GPUs with a learning rate of $1 \times 10^{-4}$ and a batch size of 32. Please see supplementary material for more details.

\subsection{Quantitative Results}
\textbf{Zero-Shot Semantic Occupancy Prediction.}
We demonstrate zero-shot semantic occupancy prediction performance of ShelfGaussian in~\cref{tab:occ3d}. Ours outperforms all existing self-supervised and weakly supervised methods, especially on vehicles (e.g., \textit{bus}, \textit{car}, \textit{trailer}), dynamic agents (e.g., \textit{bicycle}, \textit{motorcycle}, \textit{pedestrian}), and unstructured surfaces (e.g., \textit{manmade}, \textit{vegetation}), due to Gaussians' superior object-centric modeling property. 
Notably, ours surpasses the SOTA method LOcc~\cite{locc} in most metrics with only $\textbf{15\%}$ of its number of queries (i.e., $6000$ vs. $40000$). As more sensor modalities are fused, our model exhibits steady performance gains, with ShelfGaussian-LCR yielding \textbf{+6.2} IoU and \textbf{+2.71} mIoU over CO, demonstrating the effectiveness of our multi-modal Gaussian transformer.

\noindent \textbf{Zero-Shot BEV Map Segmentation.}
Existing BEV segmentation models rely heavily on ground-truth labels and require separate training for each semantic class (e.g., vehicle, pedestrian). In contrast, ours eliminates the need for human annotation and requires only a single training process. As shown in \cref{tab:bev_seg}, ShelfGaussian achieves competitive performance compared with existing SOTA methods.

\noindent \textbf{Gaussian-based Trajectory Planning.}
As shown in~\cref{tab:open_loop_planning}, Gaussian-Planner performs on par with existing top-performing methods while reducing the average CR to $\textbf{0.32\%}$, benefiting from the intrinsically object-centric characteristics of Gaussian representations. These results validate the effectiveness of our shelf-supervised Gaussians in vehicle trajectory planning and demonstrate their potential to mitigate collision risk in real-world autonomous driving.

\subsection{Ablation Study}

\noindent \textbf{Ablation on Alignment with Different VFMs.}
We perform experiments on aligning Gaussians with DINOv2~\cite{dinov2} and DINOv3~\cite{dinov3}, and using Talk2DINO~\cite{talk2dino} to query semantics. Results in~\cref{tab:ablation_vfm} lead to an empirical finding that our model with DINOv3 excels DINOv2 on some vehicles (e.g., \textit{car}, \textit{trailer}) and small long-tail classes (e.g., \textit{traffic cone}, \textit{others}) but degrades catastrophically on dynamic road agents (e.g., \textit{motorcycle}, \textit{pedestrian}) and static unstructured surfaces (e.g., \textit{sidewalk}, \textit{manmade}). This indicates that although DINOv3 provides stronger visual representations, it still faces challenges in urban driving scenarios.

\noindent
\textbf{Ablation on Shelf-Supervision Losses.}
We conduct an ablation study on different supervision losses. As revealed in~\cref{tab:ablation_loss_tab}, 3D scene-level supervision alone yields more accurate geometric and semantic understanding than 2D image-level supervision, with performance gains primarily stemming from the 3D BCE loss and the 3D feature loss, respectively. The results also show that combining 2D and 3D supervision achieves the best performance by leveraging complementary information from both domains.

\noindent \textbf{Benchmark of G2V Splatting Module.}
We benchmark our G2V spalting module against other open-source methods~\cite{gaussianformer, gausstr} in two settings: $18$k Gaussians with $1024$-dim features and $9$k Gaussians with $768$-dim features, with a target voxel number of $640$k. As shown in~\cref{tab:g2v}, our module achieves \textbf{10x} faster forward and \textbf{5x} faster backward passes than GaussianFormer~\cite{gaussianformer}, while consuming only around \textbf{40\%} of the memory. Our G2V module enables Gaussians to learn high-dimensional open-vocabulary features from 3D data with high efficiency, laying the foundation for generalizable Gaussian-based 3D scene understanding.

\subsection{Qualitative Results}
\textbf{Semantic Occupancy Prediction.}
Qualitative results of ShelfGaussian on nuScenes and custom datasets are given in~\cref{fig:qualitative_nuscenes} and~\cref{fig:qualitative_custom}, respectively. In on-road driving scenarios, as shown in~\cref{fig:qualitative_nuscenes}, ShelfGaussian predicts fine-grained occupancy with accurate semantics and detailed geometry, exhibiting high similarity to the annotated closed-vocabulary ground-truth while being open-vocabulary. Given a natural language query (e.g., \textit{double-decker bus}), our model can localize the target object and complete its shape precisely.
In other urban scenarios, ShelfGaussian also demonstrates superior open-vocabulary occupancy prediction capability. For example, as shown in~\cref{fig:qualitative_custom}, given the text query \textit{``flower shrubs"}, our model successfully recovers all flower shrub areas, highlighted in red, despite their sparse features and low texture.

\noindent \textbf{Trajectory Planning.}
\cref{fig:planning} illustrates a failure case of BEV-Planner~\cite{bevplanner} where our Gaussian-Planner successfully prevents the collision. This indicates the crucial role of object-centric Gaussian representations in reducing collision risk for safe, robust vehicle trajectory planning.

\section{Conclusion}
\label{sec:conclusion}
In this work, we present ShelfGaussian, an open-vocabulary, multi-modal, Gaussian-based 3D scene understanding framework supervised by off-the-shelf VFMs. Unlike existing methods that align 3D Gaussians with VFMs only at the 2D image level, ours unlocks the capability of foundation model alignment at the 3D scene level, achieved by a DINO-driven pseudo labeling engine and a CUDA-accelerated G2V splatting module. This 3D alignment capability significantly improves the model's zero-shot performance, narrowing the performance gap to fully-supervised methods while alleviating scalability and generalization limitations. Furthermore, our model is general, supporting a combination of multi-sensor inputs, and versatile, achieving superior performance on three downstream tasks. Notably, ShelfGaussian achieves SOTA zero-shot performance on semantic occupancy prediction on nuScenes dataset and superior in-the-wild performance on a UGV platform.

\section*{Acknowledgments}
\label{sec:acknowledgments}
Lingjun Zhao is supported by the Institute for Robotics and Intelligent Machines (IRIM) Ph.D. Fellowship at Georgia Institute of Technology.
This work used Delta GPU and DeltaAI at NCSA through allocation CIS251067 from the Advanced Cyberinfrastructure Coordination Ecosystem: Services \& Support (ACCESS) program, which is supported by U.S. National Science Foundation grants \#2138259, \#2138286, \#2138307, \#2137603, and \#2138296.

{
    \small
    \bibliographystyle{ieeenat_fullname}
    \bibliography{main}
}
\clearpage
\setcounter{page}{1}
\maketitlesupplementary


\noindent In this supplementary material, we first describe the robot hardware setup and our custom dataset collection process in~\cref{sec:supp_custom_dataset}. Then we conduct more ablation experiments in~\cref{sec:supp_ablation}. More qualitative results are provided in~\cref{sec:supp_qualitative}. The implementation of our CUDA-accelerated Gaussian-to-voxel (G2V) splatting module is detailed in~\cref{sec:supp_cuda}. Finally, we discuss the limitations and future directions in~\cref{sec:supp_limitation}.

\section{In-the-Wild Robot Experiment}
\label{sec:supp_custom_dataset}
\subsection{UGV Hardware Setup}
We illustrate the setup of our unmanned ground vehicle (UGV) in this section. A Clearpath Jackal robot is employed as our base UGV in this experiment. For collecting monocular RGB images, we use an on-board ZED 2i stereo camera. For collecting paired LiDAR point clouds, we utilize one Velodyne VLP16 LiDAR and one Hesai XT16 LiDAR. The layout of the coordinate frames of all sensors is given in~\cref{fig:ugv_setup}. ROS2~\cite{ros2} is adopted for the communication between the robot, sensors and personal computer. The intrinsics of ZED 2i camera is calibrated through built-in ZED SDK while the extrinsics of LiDARs are calibrated via open-source toolkit~\cite{Donoso2025ros2_camera_lidar_fusion}.

\begin{figure}[h] 
    \centering
    \includegraphics[width=0.8\columnwidth]{figure/supp_robot_frame.pdf}
    \caption{\textbf{Visualization of the coordinate frames of different sensors and the ego vehicle.} \textcolor{red}{Red}, \textcolor{green}{green} and \textcolor{blue}{blue} arrows denote the $x$, $y$ and $z$ axes, respectively.}
    \label{fig:ugv_setup}
\end{figure}

\subsection{Custom Dataset Collection}
\begin{figure*}[h] 
    \centering
    \includegraphics[height=5.7cm]{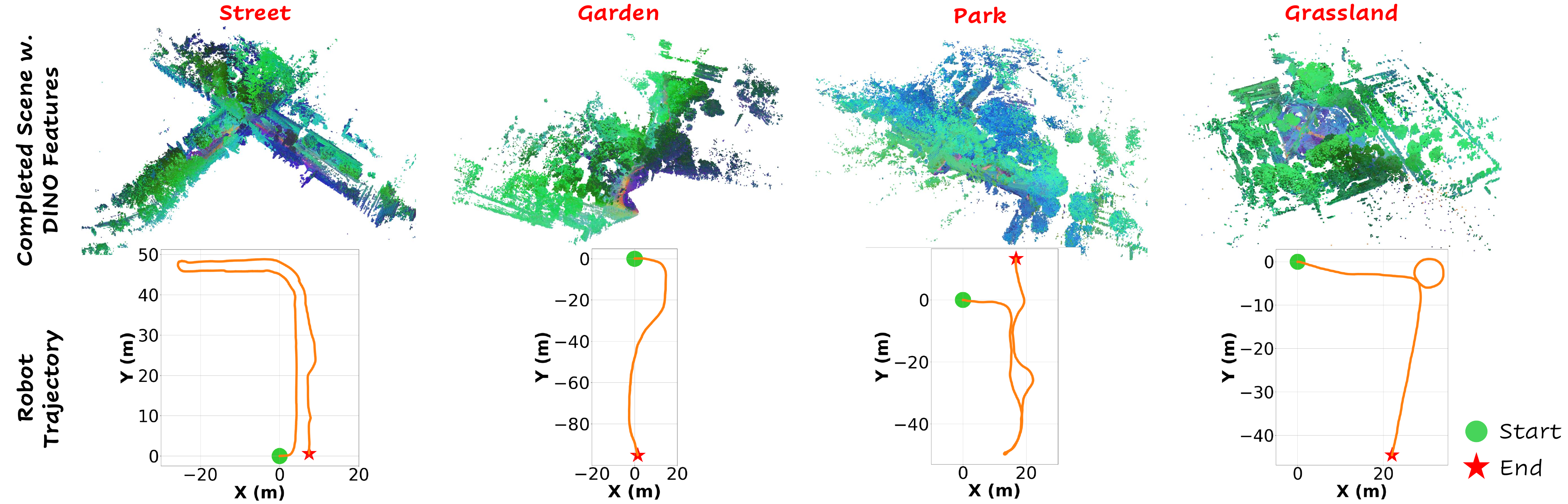}
    \caption{\textbf{Scene reconstruction results of four urban scenes.} The top row shows the completed scenes decorated with DINO features, visualized by mapping PCA components to RGB colors. The bottom row shows the robot trajectories within four urban scenes.}
    \label{fig:supp_scene_reconstruction}
\end{figure*}

By teleoperating our UGV, we collect a custom dataset in common urban scenarios. We choose four scenes: \textit{street}, \textit{park}, \textit{grassland} and \textit{garden}. We split our dataset into a 90\% subset for training and a 10\% subset for testing, resulting in $2638$ and $294$ samples, respectively. The frequency of the key frames is set to $2$Hz. We use our DINO-driven pseudo labeling engine to generate pseudo labels at key frames in a range of 
$[0\rm{m}, 20\rm{m}] \times [-10\rm{m}, 10\rm{m}] \times [-1\rm{m}, 4\rm{m}]$ with a voxel resolution of $0.2\mathrm{m}$. Each voxel label is decorated with $128$-channel compressed DINO features. We visualize the scene reconstruction in~\cref{fig:supp_scene_reconstruction}.

\section{Ablation Study}
\label{sec:supp_ablation}
We conduct more ablation experiments in this section to study the effectiveness of individual modules and justify our architectural choices.
Unless otherwise noted, all ablation studies are conducted on our model with camera-only input using the DINOv2 ViT-B/14~\cite{dinov2} model, $1000$ Gaussian queries per view, and the Talk2DINO~\cite{talk2dino} language alignment method, ensuring consistency with our main paper.

\subsection{Ablation on Number of Gaussians}
We conduct an ablation study on the number of Gaussians per view to examine its effect on model accuracy. As shown in~\cref{tab:supp_ablation_gaussian_number}, with an increased number of Gaussian queries, the final model performance improves steadily, although some metrics exhibit minor fluctuations.

\begin{table}[!h]
    \centering
    \resizebox{0.8\linewidth}{!}{
    \begin{tabular}{c|ccc|>{\columncolor{black!10}}c>{\columncolor{black!10}}c}
    
    \midrule
    \multirow{2}{*}{Mod.} & 
    \multicolumn{3}{c|}{Number of Gaussians} & ~ & ~\\
    ~ & 300 & 600 & 1000 & \multirow{-2}{*}{IoU} & \multirow{-2}{*}{mIoU} \\
    \midrule
    
    \cellcolor{white}~& \checkmark & & & 61.42 & 17.85\\
    
    \cellcolor{white}~ &  & \checkmark & & 61.07 & 18.26\\
    
    \cellcolor{white} \multirow{-3}{*}{C}  &  &  & \checkmark & 63.25 & 19.07\\
    
    \hline
    
    \end{tabular}
    }
    \caption{\textbf{Ablation on the number of Gaussian queries.}}
    \label{tab:supp_ablation_gaussian_number}
\end{table}

\subsection{Ablation on 2D and 3D Loss Weights}
To study the effects of 2D and 3D losses on the final performance, we conduct experiments with different 3D loss weights. The weights for the BCE and feature losses are empirically selected based on their magnitudes relative to the 2D losses. As shown in~\cref{tab:supp_loss_weights}, increasing 3D loss weights consistently boosts both IoU and mIoU.

\begin{table}[!h]
    \centering
    \resizebox{\linewidth}{!}{
    \begin{tabular}{c|c|cc|>{\columncolor{black!10}}c>{\columncolor{black!10}}c}
    
    \midrule
    \multirow{2}{*}{Mod.} & 
    \multirow{2}{*}{2D Loss} &
    \multicolumn{2}{c|}{3D Loss}  & ~ & ~\\
       & 
       & BCE Loss & Feat. Loss & \multirow{-2}{*}{IoU} & \multirow{-2}{*}{mIoU}
    \\
    \midrule
    \multirow{3}{*}{C} & 1.0 & 1.0 & 1.0 & 58.66 & 17.52\\
    ~ & 1.0 & 4.0 & 8.0 & 61.38 & 18.56 \\
    ~ & 1.0 & 8.0 & 16.0 & 63.25 & 19.07 \\

    \hline
    
    \end{tabular}
    }
    \caption{\textbf{Ablation on 2D and 3D loss weights.}}
    \label{tab:supp_loss_weights}
\end{table}

\subsection{Ablation on Language Alignment Methods}
Talk2DINO~\cite{talk2dino} and DINO.txt~\cite{dinotxt} are two popular approaches for aligning natural language with DINO~\cite{dinov2, dinov3} features. Talk2DINO~\cite{talk2dino} directly learns a mapping layer to transform CLIP~\cite{clip} embeddings of text queries into the DINO feature space. DINO.txt~\cite{dinotxt} learns a vision block on top of the original DINO backbone and a text encoder for raw text queries, and then aligns the encoded vision and text features through contrastive learning. We test these two methods in our framework, and the results are shown in~\cref{tab:supp_language_alignment}. With the same VFM backbone, our model using Talk2DINO~\cite{talk2dino} demonstrates better performance than using DINO.txt~\cite{dinotxt}, especially in mIoU. This indicates that directly encoding Gaussians with DINO features leads to stronger semantic and physical representations.

\begin{table}[!h]
    \centering
    \resizebox{\linewidth}{!}{
    \begin{tabular}{c|c|c|c|>{\columncolor{black!10}}c>{\columncolor{black!10}}c}
    \midrule
    Mod. & 
    Sup. &
    VFM  & Align. & IoU & mIoU\\
    \midrule
    \multirow{2}{*}{C} & \multirow{2}{*}{2D} & \multirow{2}{*}{DINOv3 ViT-L/16~\cite{dinov3}} & Talk2DINO~\cite{talk2dino} & 46.55 & 12.04 \\
    ~ & ~ & ~ & DINO.txt~\cite{dinotxt} & 47.08 & 5.71 \\ 
    \hline
    
    \end{tabular}
    }
    \caption{\textbf{Ablation on language alignment methods.} Since DINO.txt~\cite{dinotxt} only releases model weight based on DINOv3 ViT-L/16~\cite{dinov3}, we conduct ablation experiments using this. PCA~\cite{pca} is utilized to compress the feature dimension from $1024$ to $128$.}
    \label{tab:supp_language_alignment}
\end{table}

\subsection{Ablation on Feature Compression Dimensions}
To explore the influence of (PCA)~\cite{pca} feature compression on the final model performance, we conduct ablation experiments with different compressed feature dimensions. As shown in~\cref{tab:supp_pca_dim}, increasing compressed dimension generally improves the final model performance, particularly in the case of 2D and 3D joint supervision. However, the computational overhead increases as more feature channels are involved in the forward and backward passes. Considering the tradeoff between computational efficiency and model performance, we eventually choose $128$ as the final compressed feature dimension.

\begin{table}[!h]
    \centering
    \resizebox{0.8\linewidth}{!}{
    \begin{tabular}{c|cc|cc|>{\columncolor{black!10}}c>{\columncolor{black!10}}c}
    
    \midrule
    \multirow{2}{*}{Mod.} & \multicolumn{2}{c|}{Sup.} &
    \multicolumn{2}{c|}{PCA Dim.}  & ~ & ~\\
       & 2D & 3D & 2D & 3D & \multirow{-2}{*}{IoU} & \multirow{-2}{*}{mIoU}
    \\
    \midrule
    \multirow{4}{*}{C} & \checkmark & & 64 & - & 46.80 & 12.47\\
    ~ & \checkmark & \checkmark & 64 & 64 & 58.96 & 18.18\\
    ~ & \checkmark &  & 128 & - & 47.26 & 12.39\\
    ~ & \checkmark & \checkmark & 128 & 128 & 63.25 & 19.07\\
    \hline
    
    \end{tabular}
    }
    \caption{\textbf{Ablation on PCA feature compression dimensions.}}
    \vspace{-5pt}
    \label{tab:supp_pca_dim}
\end{table}

\subsection{Ablation on Trajectory Planning}
We further conduct ablation study on the effects of ego status and past trajectory on vehicle trajectory planning. As shown in~\cref{tab:supp_planning}, with the addition of vehicle state information, the planning performance is steadily enhanced. Compared to other methods using vehicle status, our Gaussian-Planner demonstrates comparable L2 error and lower collision rate, which indicates the effectiveness of object-centric Gaussian representations in reliable, collision-aware vehicle trajectory planning.
However, removing vehicle state input leads to suboptimal performance, suggesting that combining object-centric Gaussians with basic vehicle status remains beneficial and offers an avenue for further refinement.

\begin{table}[!h]
    \centering
    \resizebox{\linewidth}{!}{
    \begin{tabular}{c|cc|ccc>{\columncolor{black!10}}c|ccc>{\columncolor{black!10}}c}
    
    \midrule
    \multirow{2}{*}{Method} & \multirow{2}{*}{\makecell{Ego\\Status}} & \multirow{2}{*}{\makecell{Past\\Traj.}} &
    \multicolumn{4}{c|}{L2 (m) $\downarrow$} &
    \multicolumn{4}{c}{CR (\%) $\downarrow$} \\
       & & &
     1s & 2s & 3s & Avg. &
    1s & 2s & 3s & Avg. 
    \\
    \midrule
    ~ & & & 1.05 & 1.76 & 2.50 & 1.77 & 0.31 & 2.38 & 5.31 & 2.67 \\
    ~ & \checkmark & & 0.16 & 0.33 & 0.59 & 0.36 & 0.03 & 0.23 & 0.84 & 0.37\\
    \multirow{-3}{*}{Gaussian-Planner} & \checkmark & \checkmark & 0.16 & 0.32 & 0.58 & 0.35 & 0.00 & 0.18 & 0.78 & 0.32 \\

    \hline
    
    \end{tabular}
    }
    \caption{\textbf{Ablation on open-loop trajectory planning.} L2 (m): the mean Euclidean distance between the predicted and ground-truth trajectories. CR (\%): the percentage of the planned trajectory intersecting an obstacle. Lower is better for both metrics.}
    \label{tab:supp_planning}
\end{table}

\begin{figure*}[htp] 
    \centering
    \includegraphics[height=7cm]{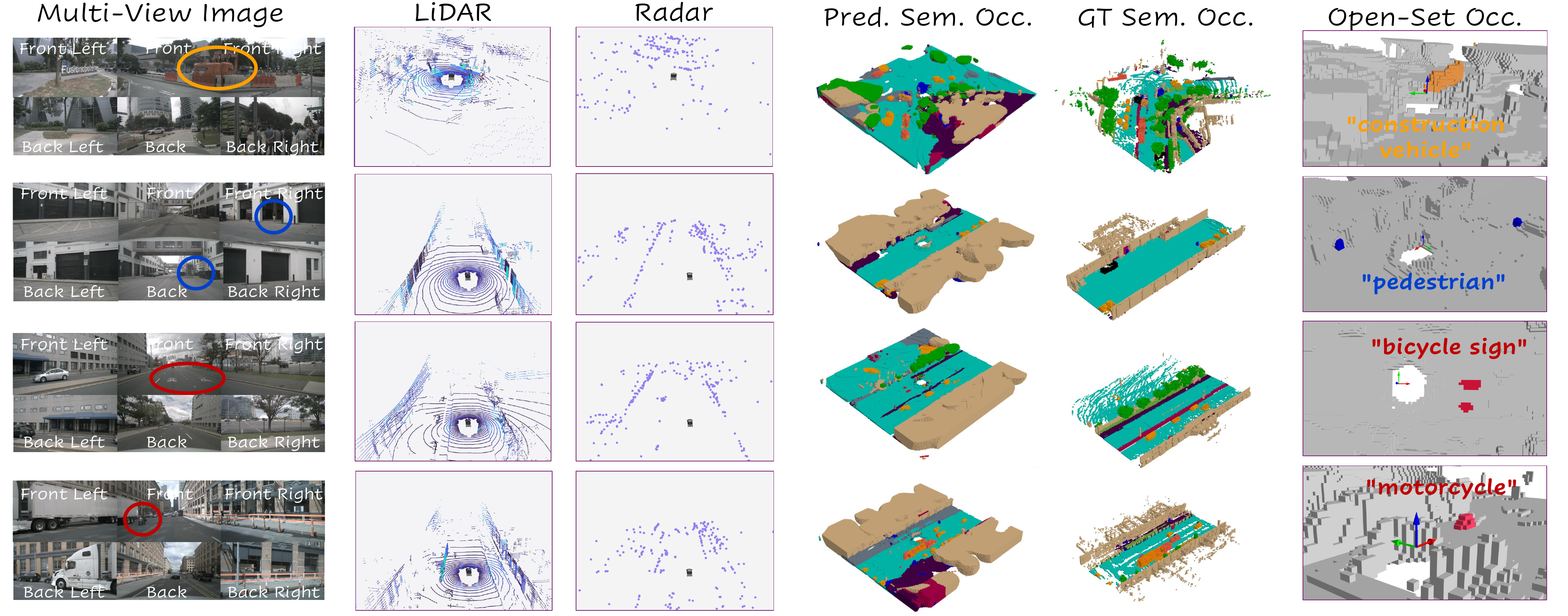}
    \vspace{-15pt}
    \caption{\textbf{Qualitative results of ShelfGaussian on nuScenes~\cite{nuscenes} dataset validation split.}}
    \vspace{-5pt}
    \label{fig:supp_qualitative_nuscenes}
\end{figure*}

\begin{figure*}[htp] 
    \centering
    \includegraphics[height=7cm]{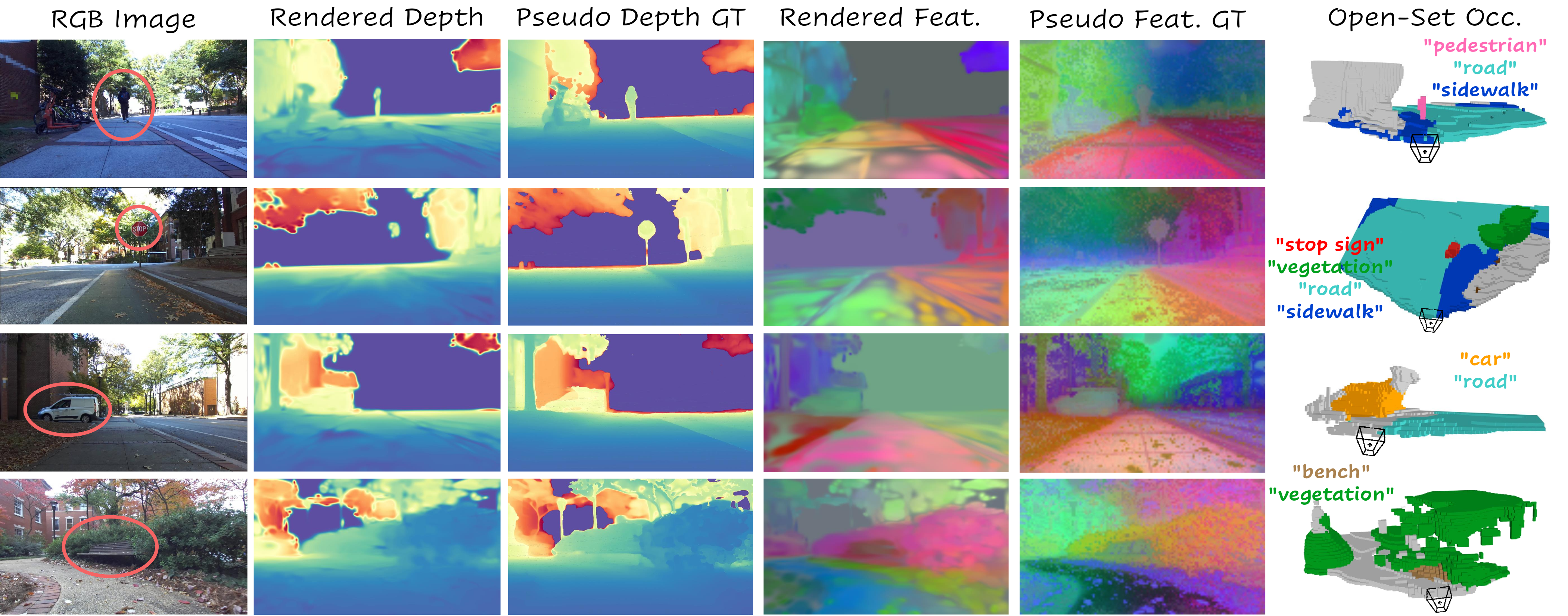}
    \vspace{-15pt}
    \caption{\textbf{Qualitative results of ShelfGaussian on our custom dataset testing split.}}
    \vspace{-10pt}
    \label{fig:supp_qualitative_custom}
\end{figure*}

\subsection{Training Efficiency}

\begin{table}[!h]
    \centering
    \resizebox{\linewidth}{!}{
    \begin{tabular}{c|c|>{\columncolor{black!10}}c>{\columncolor{black!10}}c|cc}
    
    \midrule
    Method & 
    Mod. &
    Train. Time (h) & Memory (GB)  & IoU & mIoU\\
    \midrule
    GaussTR~\cite{gausstr} & C & 22 & 20 & 44.54 & 12.27 \\
    \midrule
    \multirow{5}{*}{ShelfGaussian} & C & 25 & 15 & 63.25 & 19.07 \\
    ~ & L & 31 & 28 & 66.10 & 19.34 \\
    ~ & C+R & 31 & 28 & 62.84 & 19.42 \\
    ~ & L+C & 32 & 29 & 69.24 & 21.52 \\
    ~ & L+C+R & 32 & 29 & 69.45 & 21.78 \\
    \hline
    
    \end{tabular}
    }
    \vspace{-5pt}
    \caption{\textbf{Benchmark of training efficiency.} For a fair comparison, all experiments are conducted on 8 Nvidia L40S 48GB GPUs.}
    \vspace{-5pt}
    \label{tab:supp_training_efficiency}
\end{table}

\noindent We compare the training time and training memory usage of our model with GaussTR~\cite{gausstr}, one of the most training-efficient self-supervised methods to date. As shown in~\cref{tab:supp_training_efficiency}, our method achieves better performance than GaussTR~\cite{gausstr} with similar computational resources. As more sensor modalities are incorporated, both model performance and training cost increase accordingly.

\subsection{Gaussian-to-Voxel Splatting Efficiency}

\begin{table}[!h]
    \centering
    \resizebox{\linewidth}{!}{
    \begin{tabular}{c|ccc|ccc}
    
    \midrule
    \multirow{2}{*}{Method} & 
    \multicolumn{3}{c|}{18k Gauss. w. 1024D} &
    \multicolumn{3}{c}{9k Gauss. w. 768D} \\
       & 
     FW(s)  & BW(s) & Mem.(GB) &
    FW(s) & BW(s) & Mem.(GB) 
    \\
    \midrule
    Taichi~\cite{taichi} &- &- & Fail &- &- & Fail \\
    GaussianFlowOcc~\cite{gaussianflowocc} &- &- &OOM &- &- &OOM \\
    GaussTR~\cite{gausstr} & 14.1 & 596.7 & 22.3 & 12.4 & 446.4 & 14.8\\
    GaussianFormer~\cite{gaussianformer} & 5.4 & 13.4 & 12.9 & 1.9 & 4.2 & 7.7\\
    \midrule
    \rowcolor{black!10}
    ShelfGaussian & \textbf{0.5} & \textbf{2.7} & \textbf{4.9} & \textbf{0.2} & \textbf{1.0} & \textbf{3.7}\\

    \hline
    
    \end{tabular}
    }
    \caption{\textbf{Benchmark of G2V splatting module.} FW: forward time. BW: backward time. Mem.: memory. OOM: out of memory. Fail: exceed structural compiler limit. All the experiments are conducted on one Nvidia L40S 48GB GPU.}
    \label{tab:supp_g2v}
\end{table}

\noindent We further compare our designed G2V module with the ones used in Taichi~\cite{taichi} and GaussianFlowOcc~\cite{gaussianflowocc}, and the results are shown in~\cref{tab:supp_g2v}. Under our challenging experimental setup, Taichi~\cite{taichi} fails to complete the operation due to exceeding the structural compiler limit, while GaussianFlowOcc~\cite{gaussianflowocc} encounters out-of-memory issues. 
It is also worth mentioning that GaussianFlowOcc~\cite{gaussianflowocc} models each Gaussian’s influence using an isotropic, fixed-size cube centered at its mean position.
This approximation neglects the anisotropic covariance geometry of 3D Gaussians, leading to inaccurate computation.

\section{Qualitative Results}
\label{sec:supp_qualitative}
\subsection{nuScenes Dataset}
We provide more qualitative results of ShelfGaussian on the nuScenes~\cite{nuscenes} dataset in~\cref{fig:supp_qualitative_nuscenes}. Our model is able to predict dynamic agents (e.g., \textit{pedestrian}), long-tail objects (e.g., \textit{construction vehicle}) and novel classes (e.g., \textit{bicycle sign}) in urban driving scenarios.

\subsection{Custom Dataset}
We provide more qualitative results of ShelfGaussian on our custom dataset in~\cref{fig:supp_qualitative_custom}. By querying the model with open-vocabulary categories such as ``\textit{pedestrian}" and ``\textit{stop sign}'', it successfully localizes and completes the objects.




\section{Gaussian-to-Voxel Splatting Module}
\label{sec:supp_cuda}
This section presents a comprehensive description of the gradient computation implemented in our CUDA-accelerated G2V splatting operator. We derive the gradients with respect to Gaussians' means, covariance matrices, opacities, and high-dimensional features, respectively. 

\subsection{Preliminaries}

For a Gaussian $g$ evaluated at voxel center $x_v$, its un-normalized density
contribution to voxel $v$ is computed as:
\[
w_{g,v}
=
\alpha_g \exp\!\left(
-\frac{1}{2}(x_v-\mu_g)^\top \Sigma_g^{-1}(x_v-\mu_g)
\right),
\]
where $\mu_g \in \mathbb{R}^3$ is the mean, $\Sigma_g \in \mathbb{R}^{3\times 3}$
is the covariance matrix, and $\alpha_g$ is the opacity.
For convenience, we define:
\[
d_{g,v} = x_v - \mu_g, 
\qquad
\phi_{g,v} = \exp\!\left(
-\frac{1}{2} d_{g,v}^\top \Sigma_g^{-1} d_{g,v}
\right).
\]
The density contribution $w_{g,v}$ can be then expressed as:
\[
w_{g,v} = \alpha_g \cdot \phi_{g,v}.
\]
Each Gaussian $g$ also encodes a high-dimensional feature vector $f_g \in \mathbb{R}^C$, where $C$ is the feature dimension.
\paragraph{Voxel-wise Accumulators.}
For a fixed voxel $v$, we aggregate the contributions of all Gaussians to this voxel as:
\[
F_v = \sum_g w_{g,v}, \\
N_v = \sum_g w_{g,v}\,f_g, \\
S_v = \max(F_v,\varepsilon).
\]

\noindent Here, $F_v$ is the accumulated density at voxel $v$, $N_v$ is the density-weighted features, and $S_v$ is the clamped denominator for numerical stability. $\varepsilon$ is a constant, set as $1 \times 10^{-6}$.

\paragraph{Voxel Features.}
The final features of voxel $v$ are defined as the normalized, density-weighted
average of the Gaussian features:
\[
G_v = \frac{N_v}{S_v} \in \mathbb{R}^C.
\]

\noindent The upstream gradient with respect to the voxel features can be expressed as:
\[
\bar{G}_v = \frac{\partial \mathcal{L}}{\partial G_v} \in \mathbb{R}^C.
\]
In the following sections, we derive gradients of the loss $\mathcal{L}$ with
respect to different Gaussian parameters $(\mu_g,\Sigma_g,\alpha_g)$ and features
$f_g$ using these voxel-wise accumulators.

\subsection{Gradient w.r.t.\ Gaussian Features}

$N_v$ depends on the Gaussian features $f_g$, while $S_v$ is independent of it.
Thus, the gradients can be computed as:
\[
\frac{\partial N_v}{\partial f_g}
= w_{g,v}\, \cdot I_C,
\qquad
\frac{\partial S_v}{\partial f_g} = 0,
\]
and
\[
\frac{\partial G_v}{\partial f_g}
= \frac{1}{S_v}\frac{\partial N_v}{\partial f_g}
= \frac{w_{g,v}}{S_v} \cdot I_C,
\]
where $I_C$ denotes the identity matrix.
Applying the upstream gradient $\bar{G}_v$, we obtain the gradient with respect to the Gaussian feature $f_g$:
\[
\frac{\partial \mathcal{L}}{\partial f_g}
= \frac{\partial \mathcal{L}}{\partial G_v}\frac{\partial G_v}{\partial f_g} = \sum_v \left( \frac{w_{g,v}}{S_v} \right)\bar{G}_v.
\]

\subsection{Gradient w.r.t.\ Density Contribution}

Based on $G_v = N_v/S_v$, by applying the quotient rule, we can obtain:
\[
\frac{\partial G_v}{\partial w_{g,v}}
=
\begin{cases}
\dfrac{f_g - G_v}{S_v}, & F_v > \varepsilon. \\
\dfrac{f_g}{\varepsilon}, & F_v \le \varepsilon.
\end{cases}
\]
Therefore the gradient with respect to the density contribution $w_{g,v}$ can be calculated as:
\[
s_{g,v} =
\frac{\partial \mathcal{L}}{\partial G_v} 
\frac{\partial G_v}{\partial w_{g,v}}
=
\begin{cases}
\bar{G}_v \,\cdot\, \dfrac{f_g - G_v}{S_v}, & F_v > \varepsilon. \\
\bar{G}_v \,\cdot\, \dfrac{f_g}{\varepsilon}, & F_v \le \varepsilon.
\end{cases}
\]


\subsection{Gradient w.r.t.\ Gaussian Mean}

The density contribution of Gaussian $g$ at voxel $v$ is expressed as:
\[
w_{g,v}
= \alpha_g \exp\!\left(-\tfrac{1}{2}d_{g,v}^\top\Sigma_g^{-1}d_{g,v}\right),
\]
where $d_{g,v} = x_v - \mu_g$.
By taking the derivative of $w_{g,v}$ with respect to the Gaussian mean $\mu_g$, we have:
\[
\frac{\partial w_{g,v}}{\partial \mu_g}
= w_{g,v}\,\Sigma_g^{-1}(x_v-\mu_g).
\]
Thus, the gradient of the loss $\mathcal{L}$ with respect to $\mu_g$ can be formed as:
\[
\frac{\partial \mathcal{L}}{\partial \mu_g}
= \sum_v s_{g,v}\,w_{g,v}\,\Sigma_g^{-1}(x_v-\mu_g).
\]

\subsection{Gradient w.r.t.\ Gaussian Opacity}

Since we have $w_{g,v} = \alpha_g \cdot \phi_{g,v}$, we can compute the derivative of $w_{g,v}$ with respect to Gaussian opacity $\alpha_g$ as:
\[
\frac{\partial w_{g,v}}{\partial\alpha_g} = \phi_{g,v}
= \frac{w_{g,v}}{\alpha_g}.
\]
Therefore, the gradient of the loss $\mathcal{L}$ with respect to $\alpha_g$ can be computed as:
\[
\frac{\partial \mathcal{L}}{\partial\alpha_g}
= \sum_v s_{g,v}\,\phi_{g,v}
= \sum_v s_{g,v}\,\frac{w_{g,v}}{\alpha_g}.
\]

\subsection{Gradient w.r.t.\ Gaussian Covariance}

By taking the derivative of density contribution with respect to Gaussian covariance $\Sigma_g$, we obtain:
\[
\frac{\partial w_{g,v}}{\partial \Sigma_g}
=
\frac{1}{2} w_{g,v}\,
\Sigma_g^{-1}
d_{g,v} d_{g,v}^\top
\Sigma_g^{-1}.
\]
Thus, the gradient of the loss $\mathcal{L}$ with respect to $\Sigma_g$ can be formulated as:
\[
\frac{\partial \mathcal{L}}{\partial \Sigma_g}
=
\frac{1}{2}
\sum_v
s_{g,v}\, w_{g,v}\,
\Sigma_g^{-1}
(x_v-\mu_g)(x_v-\mu_g)^\top
\Sigma_g^{-1}.
\]



\section{Limitations and Future Work}
\label{sec:supp_limitation}
Our current 2D image-level alignment and 3D LiDAR-based pseudo labeling approach is prone to occlusion, causing ShelfGaussian to perform suboptimally on flat surfaces (e.g., \textit{drivable surface}) and small objects (e.g., \textit{barrier}). These findings suggest future directions in exploring occlusion-aware alignment and more reliable, robust 3D supervision signals. Additionally, ShelfGaussian only supports learning 3D Gaussians on a per-view basis, which overlooks the interactions of Gaussians across different camera views. Future work could explore learning Gaussians directly in world space to eliminate view-specific restrictions.
Lastly, a more extensive evaluation of our Gaussian-based trajectory planner in more challenging driving scenarios such as WOD-E2E~\cite{wod} and on closed-loop benchmarks such as Bench2Drive~\cite{bench2drive} is an essential step toward real-world deployment.


\end{document}